\RequirePackage{fix-cm}

\documentclass[smallextended]{svjour3}
\usepackage{amsmath,amsfonts}
\usepackage{algorithmic}
\usepackage{algorithm}
\usepackage{array}
\usepackage{multirow}
\usepackage[caption=false,font=normalsize,labelfont=sf,textfont=sf]{subfig}
\usepackage{textcomp}
\usepackage{stfloats}
\usepackage{booktabs}
\usepackage{url}
\usepackage{verbatim}
\usepackage{graphicx}
\usepackage{cite}
\begin{document}

\title{XAI and Few-shot-based Hybrid Classification Model for
Plant Leaf Disease Prognosis}

\author{Diana Susan Joseph \and Pranav M Pawar$^*$ \and Raja Muthalagu \and Mithun Mukharjee
}

\institute{Diana Susan Joseph, Pranav M Pawar$^*$, Raja Muthalagu, Mithun Mukherjee, 
	\newline Department of Computer Science\\
	Birla Institute of Technology and Science Pilani, Dubai Campus, Dubai, UAE\\
	\email{p20200010@dubai.bits-pilani.ac.in, pranav@dubai.bits-pilani.ac.in, raja.m@dubai.bits-pilani.ac.in, mithun@dubai.bits-pilani.ac.in
	\newline $^*$Corresponding Author: Pranav M Pawar,pranav@dubai.bits-pilani.ac.in }           %  \\
}
\markboth{IEEE Transactions on XXX,~Vol.~XX, No.~X, June~2025}%
{Joseph \MakeLowercase{\textit{et al.}}: XAI and Few-shot Based Hybrid Classification Model for Crop Disease Prognosis}
\maketitle
%%%%%%%%%%%%%%%%%%%%%%%%%%%%%%%%%%%%%%%%%%%%%%%%%%%
%\markboth{IEEE Transactions on <Your Society>, Vol. XX, No. X, Month 2025}%
%{Joseph \MakeLowercase{\textit{et al.}}: <Short Title>}
%\author{IEEE Publication Technology,~\IEEEmembership{Staff,~IEEE,}
        % <-this % stops a space
%\thanks{This paper was produced by the IEEE Publication Technology Group. They are in Piscataway, NJ.}% <-this % stops a space
%\thanks{Manuscript received April 19, 2021; revised August 16, 2021.}}


%\IEEEpubid{0000--0000/00\$00.00~\copyright~2025 IEEE}

\begin{abstract}
Performing a timely and accurate identification of crop diseases is vital to maintain agricultural productivity and food security. The current work presents a hybrid few-shot learning model that integrates Explainable Artificial Intelligence (XAI) and Few-Shot Learning (FSL) to address the challenge of identifying and classifying the stages of disease of the diseases of maize, rice, and wheat leaves under limited annotated data conditions. The proposed model integrates Siamese and Prototypical Networks within an episodic training paradigm to effectively learn discriminative disease features from a few examples. To ensure model transparency and trustworthiness, Gradient-weighted Class Activation Mapping (Grad-CAM) is employed for visualizing key decision regions in the leaf images, offering interpretable insights into the classification process. Experimental evaluations on custom few-shot datasets developed in the study prove that the model consistently achieves high accuracy, precision, recall, and F1-scores, frequently exceeding 92\% across various disease stages. Comparative analyses against baseline FSL models further confirm the superior performance and explainability of the proposed approach. The framework offers a promising solution for real-world, data-constrained agricultural disease monitoring applications.

\end{abstract}

\section{Introduction}
One of the difficult tasks for farmers worldwide is the identification and management of plant diseases \cite{arsenovic2019solving, joseph2023intelligent, joseph2024plant}. If the diseases are not identified at the right time, it can cause a threat to food security, a decrease in yield, and affect the quality of production \cite{nagaraju2020systematic}. Similar to human health, plant health is susceptible to diseases caused by viruses and bacteria, underscoring the importance of diligent plant care and precise disease diagnosis \cite{oad2024plant, mindhe2020plant}. 
Traditional methods of plant disease diagnosis, which rely on visual assessments by experts, are often time-consuming and prone to inaccuracies

In recent years, automated approaches such as machine learning \cite{ramesh2018plant, ahmed2023plant}, deep learning \cite{ahmad2023survey, ferentinos2018deep, joseph2024mobile}, and few-shot learning \cite{rezaei2024plant, sun2024few, li2021semi, argueso2020few, lin2022few, uskaner2024efficient, joseph2024rice} have been increasingly employed to enable faster and more accurate identification and classification of plant diseases \cite{cap2020leafgan, qadri2024advances}. Most of the works that used deep learning \cite{panchal2023image, mohameth2020plant, chowdhury2021automatic} for disease diagnosis have proved to give promising results, in cases where the dataset where of good quality and there was a sufficient number of samples available for both training and testing

In situations where a newly emerging plant disease is identified and only a limited number of samples are available, traditional machine learning and deep learning models often struggle to achieve satisfactory performance due to data scarcity. Few-shot learning offers a compelling alternative, as it is capable of learning from as few as one to five samples. Recent studies have demonstrated the effectiveness of few-shot learning methods in such low-data scenarios, showing promising results that are often comparable to those achieved by conventional deep learning models. Few-shot learning methods are particularly valuable for detecting the stages of plant diseases in scenarios where only a limited number of annotated samples are available

The concept of Explainable Artificial Intelligence (XAI) has gained significant traction in recent research aimed at enhancing the interpretability of models developed for plant disease diagnosis. In fact, XAI techniques enable visualization of the specific regions within an image that a model focuses on when making predictions, thereby improving transparency and trust in the model’s decision-making process. This not only increases the reliability of the results for end users, such as agronomists and farmers, but also facilitates model debugging and refinement. In the present work, various state-of-the-art XAI methods—such as Grad-CAM \cite{selvaraju2017grad, gopalan2025corn}, Grad-CAM++ \cite{preotee2024approach}, and Eigen-CAM \cite{muhammad2020eigen, jahin2025soybean} are employed to highlight the discriminative features used by the model for disease classification

\subsection{Motivation}
Staple crops such as rice, wheat, and maize are considered to be the foods that provide humans with the required energy as they contain carbohydrates, minerals, and proteins that form a balanced diet. Therefore, the important diseases affecting these plants are considered in our study. Four diseases affecting each of these crops are considered. Two bacterial and two fungal diseases are considered for Rice, and for maize and wheat, four fungal diseases are considered each. The diseases considered for maize are CommonRust \cite{pillay2021quantifying}, Northern Leaf Blight \cite{razzaq2019study}, Grey Leaf Spot \cite{dhami2015review}, and Southern Rust \cite{sun2021southern}. The diseases considered for wheat were LeafRust \cite{prasad2020progress}, Powdery Mildew \cite{basandrai2017powdery}, Stripe Rust \cite{schwessinger2017fundamental}, and Tanspot \cite{laribi2024tan}. The diseases considered for rice are Rice Bacterial Blight \cite{sanya2022review}, Rice Blast \cite{fernandez2023phantom}, Rice Bacterial Leaf Streak \cite{chen2021approach}, and Brown Spot \cite{surendhar2022status}. The stages of the disease considered in the work are early, advancing, and severe

\subsection{Contributions}
The main contributions are summarized as follows:
\begin{enumerate}
    \item Few-shot datasets are developed from the datsets in \cite{joseph2024real} for the disease stages of diseases of rice, wheat, and maize crops considered in the study. These datasets are useful for episodic training used in few-shot methods.
    \item We test the datasets on various fined tuned few-shot learning models like Siamese Model \cite{li2022survey}, Prototypical Model \cite{snell2017prototypical}, Relational Model \cite{sung2018learning}, and Maching Networks \cite{vinyals2016matching} and the performance is recorded.
    \item We introduce a hybrid few-shot learning model for disease stage classification for the rice, wheat, and maize crops.
    \item Finally, visualization of the features concentrated by the model for disease stages classification is performed by the Class Activation Mapping (CAM) Methods.
\end{enumerate}

%
%The proposed hybrid model was able to obtain an accuracy, precision,  F1-score, and recall of above 92\% for all the disease stages of the diseases considered for the rice, wheat, and maize crops. The few-shot datasets developed for the three crop types, when evaluated using the proposed Siamese model, yielded accuracy, precision, recall, and F1-scores exceeding 70\% for rice, maize, and wheat
%
%The prototypical network developed in this study was able to obtain an accuracy of around 94\% for the wheat leaf dataset, above 95\% for the rice leaf dataset, and above 95\% for the maize leaf dataset. The Relational model achieved validation accuracy, precision, recall, and F1-scores above 79\% across all disease stages in the few-shot wheat, rice, and maize leaves datasets
%
%In the case of Matching Networks, the model was able to acquire a validation accuracy of only a maximum of 77\% for the disease stages of maize plants, a maximum of 79\% for disease stages of rice plants, and a maximum of only 72\% for disease stages of wheat plants

\subsection{Organization of the Paper}
The rest of the paper is organized as follows. Section II deals with related works in the field of disease diagnosis in plants, including methods that use Few-shot Learning (FSL), XAI that improves the explainability of the models, and Convolutional Neural Networks (CNN) \cite{HussainT22, sy, mahadik2023efficient}. Section III deals with materials and methods used in the current work, like the dataset developed and used, the details about the proposed methodology for each of the FSL methods used, and the methodology of the proposed hybrid few-shot learning model. Section IV deals with the results of FSL models and the proposed few-shot hybrid model on the few-shot datasets developed for rice, wheat, and maize crops. Section V gives the conclusion.

\section{Related Work}
\subsection{Meta-learning and Feature Attention Approaches}
A novel FSL method was proposed in \cite{rezaei2024plant} as a solution to the issue of the availability of a smaller amount of data for the identification of illness in plants. The key concept used in their work to develop an FSL method was based on meta-learning, pre-training, and fine-tuning. Along with these methods, the feature attention technique was also included in their method, which helps to identify the relevant parts of the image apart from complex backgrounds. Two feature extractors were used in their research.

A method based on meta-learning is proposed in \cite{chen2021meta} that detects unknown samples when only a few labeled examples are available. The proposed method in the research is based on conditional neural adaptive processes and feature matching. A dataset containing 26 species of plants with 60 diseases was contributed by the researchers to train the proposed method. Similarly, \cite{mu2024few} proposed a method based on supervised contrastive learning and meta-learning to detect plant diseases when only a few samples are available. The use of label information enabled the proposed algorithm to obtain better accuracy with a small batch size; therefore only fewer GPU resources were required.

\subsection{Siamese and Metric-Learning Based Models}
The research conducted in \cite{argueso2020few} presented that it is possible to achieve good results with small training samples by using few-shot learning methods like Siamese Networks with Triplet loss. The images from the Plant Village (PV) dataset \cite{hughes2015open} were used in their work. The dataset was split into source and domain classes to conduct the experiments. A pre-trained model was used in the research conducted in \cite{uskaner2024efficient} for the extraction of features from the images, and later refined using the PlantCLEF dataset. Support Vector Machine (SVM) was used for classification after the proposed FSL method was trained on the PV dataset.

The research conducted by \cite{saad2024plant} proposed a method that is well-suited for real-time applications with a processing time of only 5 ms. The proposed method is based on Siamese Networks, and a dataset containing five plants was used. The research includes augmentation techniques and a unique segmentation method. Evaluating the proposed method with specific performance metrics shows that it outperforms other ML algorithms with only a few training samples. Similarly, the work done by \cite{lin2022few} demonstrates that FSL techniques can act as a potential solution for plant disease diagnosis in low-data scenarios. The proposed methodology was based on multi-scale fusion for better feature representations and channel attention for feature learning.

The severity of strawberry leaf scorch disease was estimated in \cite{pan2022automatic}, where a two-stage method based on Faster R-CNN and a Siamese network was used. In stage one, the leaf patches are identified to form a new dataset. In the second stage, the severity of the disease is determined by training on the dataset formed. The experimental results showed that the proposed system could present good performance compared to other models.

\subsection{Semi-Supervised and Explainable FSL Approaches}
A semi-supervised FSL approach was proposed in \cite{li2021semi}. The dataset used was the PlantVillage dataset, split into three modes. Many experiments were conducted to analyze the accuracy of the proposed system. The system contained two methods: single semi-supervised and iterative. The experiments were repeated ten times for each group.
Explainable AI (XAI) techniques have also been incorporated into FSL research. XAI methods such as LIME and SHAP were used in \cite{paul2024study} to make predictions more trustworthy. Potato leaf disease diagnosis was carried out using a deep ensemble model that achieved an accuracy above 99\%. Similarly, \cite{arvind2021deep} experimented with multi-class tomato leaf disease datasets to diagnose plant diseases and identified EfficientNet B5 as the best-performing model. To increase model interpretability, XAI techniques like LIME and Grad-CAM were applied to the outputs and evaluated using YOLO v4 \cite{aldakheel2024detection,jiang2022review}.

The research conducted by \cite{thakur2022explainable} proposed a lightweight CNN and Vision Transformer hybrid model for agricultural services, evaluated on five publicly available datasets. A hybrid model was also proposed in \cite{ghosh2023recognition} to detect sunflower diseases. The model was developed using transfer learning and CNN. Among eight tested models, VGG-19 combined with CNN performed best.

\section{Materials and Methods}
\subsection{Dataset Used}
Few-shot datasets were formed from the datasets formed in \cite{joseph2024real}. The diseases affecting the rice, wheat, and maize crops were used to form a few-shot dataset for each disease considered in the work \cite{joseph2024real}. For example, a few-shot dataset for CommonRust disease stages was developed in the current work, and similarly for other diseases considered in \cite{joseph2024real}. The disease stages of the diseases considered in \cite{joseph2024real} were grouped into different classes with the help of an experienced plant pathologist. The developed dataset included four classes, namely early, advancing, severe, and healthy stages of a particular disease

\begin{figure*}[]
\center
\resizebox{12cm}{!}{% 
\includegraphics[height=5cm, width=12cm]{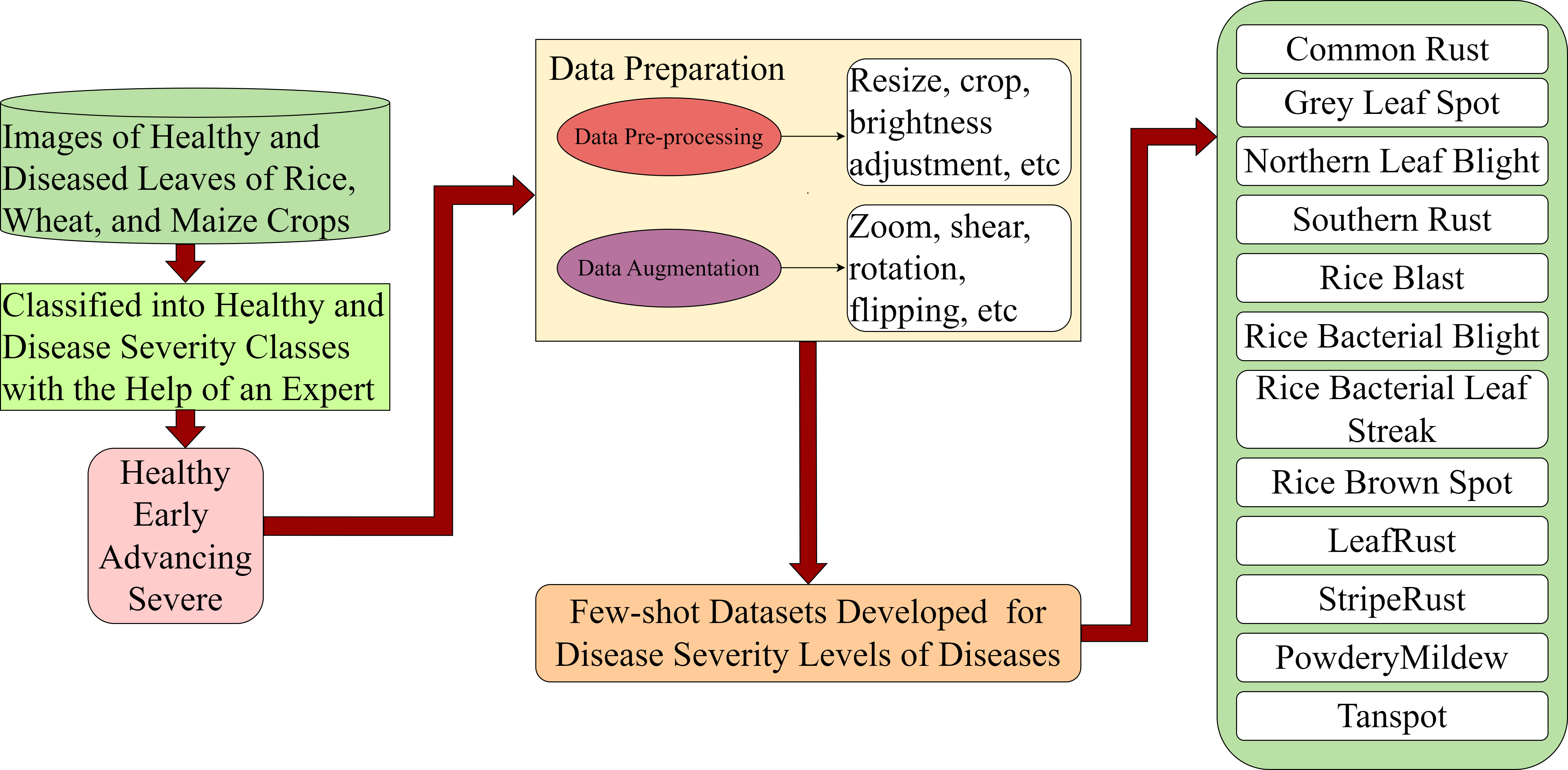}
}
\captionsetup{justification=centering}
\caption{Data Preparation for the Development of Few-shot Datasets}
\label{fig 1}       % Give a unique label
\end{figure*}

Data pre-processing and data augmentation steps were applied to the images in the dataset after organizing them into various disease severity stages. Some of the pre-processing steps applied were resizing, cropping, brightness adjustment, etc. The data augmentation steps applied were similar to those applied in the research work \cite{joseph2024real}; some of the augmentation steps included zoom, shear, rotation, flipping the images, etc. The augmentation techniques helped to increase the size of the few-shot datasets and also to experiment with different samples of data. The images in each class of the few-shot datasets were increased to a maximum of 5000 images and then divided into training, validation, and testing sets for a few-shot task formulation during the experimentations. The process of data preparation and the formation of datasets for the crops rice, wheat, and maize are shown in Fig.~\ref{fig 1}
\subsection{Proposed Methodology}
The few-shot datasets formed for each of the diseases were trained and evaluated using customized few-shot learning techniques like a Siamese model, relational Network, Matching Network, and Prototypical network. Finally, a hybrid model was developed by selecting the few-shot techniques that performed best on these datasets. A brief explanation of the architectures used for each of the few-shot techniques used for training and evaluating the few-shot datasets is provided as follows.
\begin{itemize}
    \item Siamese Network: The few-shot datasets formed for all three crops were trained and evaluated using the proposed Siamese model.  The evaluation of the datasets is made based on distance comparison between the images. The main aim of the proposed Siamese model was to learn the feature embeddings for image triplets (anchor, positive, and negative) such that the distance between the anchor and positive samples is less when compared to the anchor and negative samples. A custom CNN is used for feature extraction with two convolution layers followed by batch normalization and max pooling. The output of the CNN model is flattened and passed through a fully connected layer that outputs a 64-dimensional embedding vector. The loss function used is Triplet Loss; the loss is computed as the difference between positive and negative distances among the samples. The loss encourages the model to increase positive distance and reduce negative distance. For each batch of training, the images are passed through the Siamese network to get their embeddings, the triplet loss is calculated, and backpropagation is performed to optimize the model parameters. The model is trained with a batch size of 32 for 100 epochs, effectively exposing it to thousands of varied triplets. For each epoch, a new set of anchor–positive–negative triplets is constructed by randomly selecting images from the same and different classes. After training, the model is evaluated using the query dataset by comparing distances between embeddings. A query is considered correctly classified if its positive pair is closer than its negative pair in the embedding space. Evaluation metrics such as accuracy, precision, recall, and F1-score are reported.
 \item Relational Network: The proposed Relation Network is designed for few-shot image classification, where the goal is to determine whether a query image belongs to the same class as any of the support images. Instead of directly predicting class labels, the model computes a relation score between each query-support pair, allowing the system to adapt quickly to unseen classes at test time. The main components included are data preparation, episodic sampling, the network architecture, episodic training, and evaluation. The architecture consists of two key components: Feature Extractor and Relation Module. A ResNet-18 model \cite{he2016deep} (excluding the final fully connected layers) is used to extract deep feature maps from both support and query images. Each image is represented as a feature map $z \in \mathbb{R}^{C \times H \times W}$, where $C$ is the number of channels and $H \times W$ is the spatial resolution. In the relation module, for each support-query pair, the extracted feature maps are concatenated along the channel dimension to form a joint representation. This combined feature map is passed through a series of convolutional layers, followed by global average pooling and a fully connected layer to output a relation score.
Let $z_{\text{support}}, z_{\text{query}} \in \mathbb{R}^{C \times H \times W}$ denote the feature maps of a support and a query image, respectively. The concatenated tensor  is defined as~\cite{sung2018learning}
\begin{equation}
    z_{\text{combined}} = \text{Concat}(z_{\text{support}}, z_{\text{query}}) \in \mathbb{R}^{2C \times H \times W}\:.
\end{equation}

This combined tensor is processed by the relation module, and the final relation score, denoted as $r$  \cite{sung2018learning}\cite{chen2019closer} is computed as
\begin{equation}
    r = \sigma \left( W \cdot \text{Flatten}(\textsf{GAP}(f(z_{\text{combined}}))) + b \right)\:,
\end{equation}
where $f$ denotes the convolutional layers in the relation module, $\textsf{GAP}(\cdot)$ is the global average pooling operation,
$W$ and $b$ are the weight matrix and bias term of the final linear layer,
and $\sigma$ is the sigmoid activation function.
During training, the relation scores are compared to binary labels using the Binary Cross-Entropy (BCE) loss. For each support-query pair, we can write
\[
    y = 
    \begin{cases}
        1, & \text{if query and support are from the same class} \\
        0, & \text{otherwise}\:.
    \end{cases}
\]
It is important to note that the BCE loss encourages high relation scores for similar pairs and low scores for dissimilar ones, enabling the network to learn meaningful similarity metrics. The training is conducted for multiple episodes and early stopping is used to prevent overfitting. The model is optimized using the Adam optimizer. Each episode consists of four distinct classes, five support and query images each. Model evaluation is performed using separate few-shot episodes on the test set. For each query image, the model selects the class of the most similar support image. Evaluation metrics include accuracy, precision, recall and F1-score.
\item Matching Networks: The Network model is designed for few-shot image classification using episodic training, where the network learns to generalize from limited labeled examples from the custom datasets. Each training episode is constructed to develop a few-shot task and contains a support set with a maximum of only thirty samples per class used for classification and a query set with a maximum of only thirty samples per class used for evaluation and loss computation. This episodic structure is implemented by a custom TaskSampler that ensures consistent sampling of $N$-way $K$-shot tasks during training and testing phases. The core of the architecture is a pre-trained ResNet-18 model, which acts as the backbone for the feature extraction. The final fully connected layer of ResNet-18 is replaced with an identity layer to output raw feature embeddings:
    %\begin{equation}
\[
\text{ResNet-18}: \quad \text{fc} \rightarrow \text{Identity}
\]
%\end{equation}
These embeddings are then passed through a custom linear layer to map them to a feature space suitable for classification as in Eq. (3):
\begin{equation}
y = \text{FC}(\text{ResNet}(i))
\end{equation}
Support and query images are normalized using L2 normalization as in Eq. (4) \cite{vinyals2016matching}\cite{zhe2019directional} to ensure that the feature vectors lie on the unit hypersphere:
\begin{equation}
\hat{m}_s = \frac{m_s}{\|m_s\|_2}, \quad \hat{m}_q = \frac{m_q}{\|m_q\|_2}\:.
\end{equation}
Later, the similarity between support and query embeddings is calculated using cosine similarity as in Eq. (5) \cite{vinyals2016matching}, which reduces to a dot product since all embeddings are L2-normalized:
\begin{equation}
\texttt{similarity}(\hat{m}_q, \hat{m}_s) = \hat{m}_q \cdot \hat{m}_s
\end{equation}

This forms a similarity matrix $k$, where each element $k_{xy}$ represents the similarity between the $x$-th query and $y$-th support embedding as in Eq. (6) \cite{snell2017prototypical} :
\begin{equation}
k_{xy} = \hat{m}_q^x \cdot \hat{m}_s^y
\end{equation}

An attention mechanism is applied over the similarity matrix to compute the relevance of each support sample for each query as in Eq. (7) \cite{vinyals2016matching}:
\begin{equation}
a_{xy} = \frac{\exp(k_{xy})}{\sum_r \exp(k_{xz})}
\end{equation}
Support labels are one-hot encoded and combined with the attention matrix to produce the final classification scores for each query image for prediction by attention-weighted labels as in Eq. (8):
\begin{equation}
\texttt{score}(q) = A \cdot Y_s
\end{equation}
Where $A$ is the attention matrix derived from similarities, $Y_s$ is the one-hot encoded label matrix for support images. The query image is classified based on the label with the highest score. The model is trained for 2000 episodes, using 4-way, 30-shot, 30-query tasks per episode. Each episode samples 4 classes, and 60 images per class (30 for support, 30 for query). The training loss is computed using cross-entropy between predicted scores and true query labels, and model parameters are updated by the Adam optimizer. The trained model is evaluated on 100 few-shot tasks from the validation set using the same episodic setup. Performance is reported using accuracy, precision, recall, and F1-score, computed using the \texttt{precision\_recall\_fscore\_support} function from \texttt{scikit-learn} \cite{pedregosa2011scikit}.
     \item Prototypical Network: The proposed prototypical network model is trained to classify unseen classes with only a few labeled examples per class. The key components involved are data preparation, task sampling for episodic learning, network architecture, training loop, and evaluation metrics. During training, the few-shot tasks are formulated using the episodic training paradigm, where each episode includes four classes from the dataset. Additionally, ten query samples are used per class for evaluation within each episode. The tasks are generated dynamically during training, validation, and testing using a custom task sampler. The model architecture mainly includes the feature encoder and the network. The feature encoder is a modified version of the ResNet-18 model, trained on the ImageNet dataset, and it is used as the backbone of the network. The final fully connected layer is removed, and a new layer is added to map extracted features into a 64-dimensional embedding space. The encoder is shared across support and query samples. The proposed prototypical network computes class prototypes as the mean of embedded support examples for each class. Given an embedded query image, classification is performed based on the negative Euclidean distance to each class prototype. This distance metric is the prediction score, and the class with the smallest distance is selected. The Adam optimizer is used, and the model is trained for a maximum of 3000 training episodes, and to avoid overfitting, early stopping is applied with a patience threshold of 20 evaluations. A learning rate of $1 \times 10^{-4} $ is used. The model achieving the highest validation accuracy is saved and used for inference. The performance of the model is evaluated using several performance metrics like accuracy, precision, recall, and F1-score. 
     \end{itemize}
\subsection{Proposed Hybrid Siamese-ProtoNet}
      The primary objective of this work was to develop a robust few-shot classification model capable of effectively learning from limited labeled samples per class, by leveraging a hybrid architecture combining Siamese Networks and Prototypical Networks. Initially, extensive experimentation was conducted using a combination of Relational Networks, Siamese Networks, and Prototypical Networks to form a hybrid model
      
       The models were selected because, from the experiments conducted and from the results shown in the tables, it can be understood that these models performed relatively well on the few-shot datasets developed for rice, wheat, and maize crops. The approach failed to give good performance, after several experimentations also the validation accuracy, precision, recall, and F1-score did not improve from 25\%. Later, the relational model was removed, and only the siamese and the prototypical networks were used to build the hybrid model
       
       The hyperparameters used to develop the hybrid model are summarized in Table ~\ref{tab:Hyperparameters}. The concept of similarity-based learning of the Siamese model and prototype-based classification in the embedding space of the prototypical model is used to form the hybrid model. The ResNet-18 is used as the feature extraction backbone in the proposed model
       
        The last two layers of the fine-tuned model are truncated and are followed by global average pooling and a fully connected layer to produce fixed-dimensional embeddings. Let $f_{\theta}(\cdot)$ denote the encoder network with parameters $\theta$. Given an input image \textit{s}, the feature representation is expressed as
         \begin{equation}
         v=f_{\theta}(s) \in R^{c}\:,
     \end{equation}
     where \textit{c} is the dimensionality of the feature vector.
     The hybrid siamese-protonet model includes mainly three components:
     \begin{enumerate}
         \item \textbf{The Prototypical Component}:  The prototypical network component of the hybrid model proposed follows the method similar to \cite{snell2017prototypical}. For a support set $S$ with $N$ number of classes and $E$ examples per class, the prototype $p_n$ for each class $n \in \{1,\ldots, N\}$ is computed as the mean of embedded support vectors of that class as~\cite{snell2017prototypical}
     \begin{equation}
      p_{n}=\frac{1}{E}\sum_{i:y_{i}=n}f_{\theta}(x_{i})  \:. 
     \end{equation}
     For each query image \textit{$x_{q}$}, the distance to each prototype is computed using Euclidean distance as~\cite{snell2017prototypical}
     \begin{equation}
     d(v_{q} ,p_{n})=||f_{\theta}(x_{q})-p_{n}||_{2}^{2}  \:.
     \end{equation}
   Classification is performed by assigning \textit{$x_{q}$} to the class of the nearest prototype as
   \begin{equation}
   y_{q}= \arg \min d(v_{q}, p_{n})\:.
   \end{equation}
   \item \textbf{The Siamese Component}: The Siamese network component of the hybrid model is similar to the work in \cite{koch2015siamese}. The Siamese Network component is added by training the encoder to produce embeddings such that samples of the same class are closer in the embedding space, and those from different classes are farther apart. A distance-based contrastive loss can be used, but in the proposed hybrid model, the Euclidean distance is implicitly enforced through the prototype-based classification. The encoder produces class-discriminative embeddings, similar to the Siamese model's intent.
   \item \textbf{Hybrid Fusion}: The proposed  Hybrid Siamese-ProtoNet model uses a single shared encoder trained end-to-end via a cross-entropy loss on the negative Euclidean distances between the query embeddings and class prototypes instead of using an ensemble or separate modules. We can write softmax over distances for prediction as~\cite{snell2017prototypical}
   \begin{equation}
       z(y_{q}=n|x_{q},s)=\frac{\exp(-d(f_{\theta} (x_{q}),p_{n}))}{\sum_{n'}\exp(-d(f_{\theta} (x_{q}),p_{n'}))}\:.
   \end{equation}
   The optimized standard cross-entropy loss of the model can be expressed as~\cite{snell2017prototypical}
   \begin{equation}
       L=-\sum_{q}\log z(y_{q}=n|x_{q},s)\:.
   \end{equation}
   The loss is differentiable and allows for end-to-end training of the encoder using both support and query sets. Overall the architecture implemented effectively integrates prototype-based class abstraction with the embedding similarity principle of Siamese networks. The shared encoder learns a generalizable feature space from a small number of examples per class, while the prototype mechanism allows for fast inference by measuring distance to class centroids
   
   The proposed hybrid model is trained using the concept of episodic training, in which every episode includes 5 samples in the support set and 10 samples in the query set for training. The model is trained episodically, where in each episode N classes are randomly selected, K support and Q query samples are drawn per class, a forward pass is performed to compute distances between query features and class prototypes, and Cross-entropy loss is minimized using the true labels of the query samples. An early stopping mechanism monitors validation accuracy with a patience of 10 epochs to prevent overfitting.
   \item \textbf{Explainability via CAM}: To interpret the decision-making process of the encoder, CAM techniques are employed, including GradCAM, GradCAM++, and EigenCAM. For example, if GradCam++ is selected for interpreting the decision-making process of the encoder. Given a query image \textit{f}, CAM visualizations are generated for each class $d $ by computing weighted gradients of the target class score $g^{d}$ with respect to the last convolutional layer of the encoder. This heatmap which is overlaid on the image to highlight important regions can be computed as~\cite{chattopadhay2018grad}.
   \begin{equation}
     \text{CAM}_{d}(f)=\text{ReLU} \sum_{r}\alpha^{d}_{r}\,B^{r} \:,
   \end{equation}
   where $B^r$ is the $r$th feature map and $\alpha^d_r$ is the weight computed from the corresponding gradient. The CAM wrapper uses the hybrid few-shot model as a fixed encoder and overlays visualizations for each class on query images, helping assess which image regions contribute most to the model’s classification decisions. The eigenvalues of the weight matrix are calculated by the EigenCAM. Afterward, the output of a particular class using the EigenCAM can be computed as~\cite{bany2021eigen}. 
   \begin{equation}
       \text{EigenCAM}=\lambda_{j} \sum^{s}_{j}w_{j}F_{j}\:,
   \end{equation}
where $\lambda_{j}$ is the $j$th eigenvalue of the weight matrix $w_{j}$ and $s$ is the number of elements in the feature vector $F$.
\end{enumerate} 
% \usepackage{booktabs}
% \usepackage{graphicx}

\begin{figure*}[]
\center
\resizebox{16cm}{!}{% 
\includegraphics[height=7cm, width=12cm]{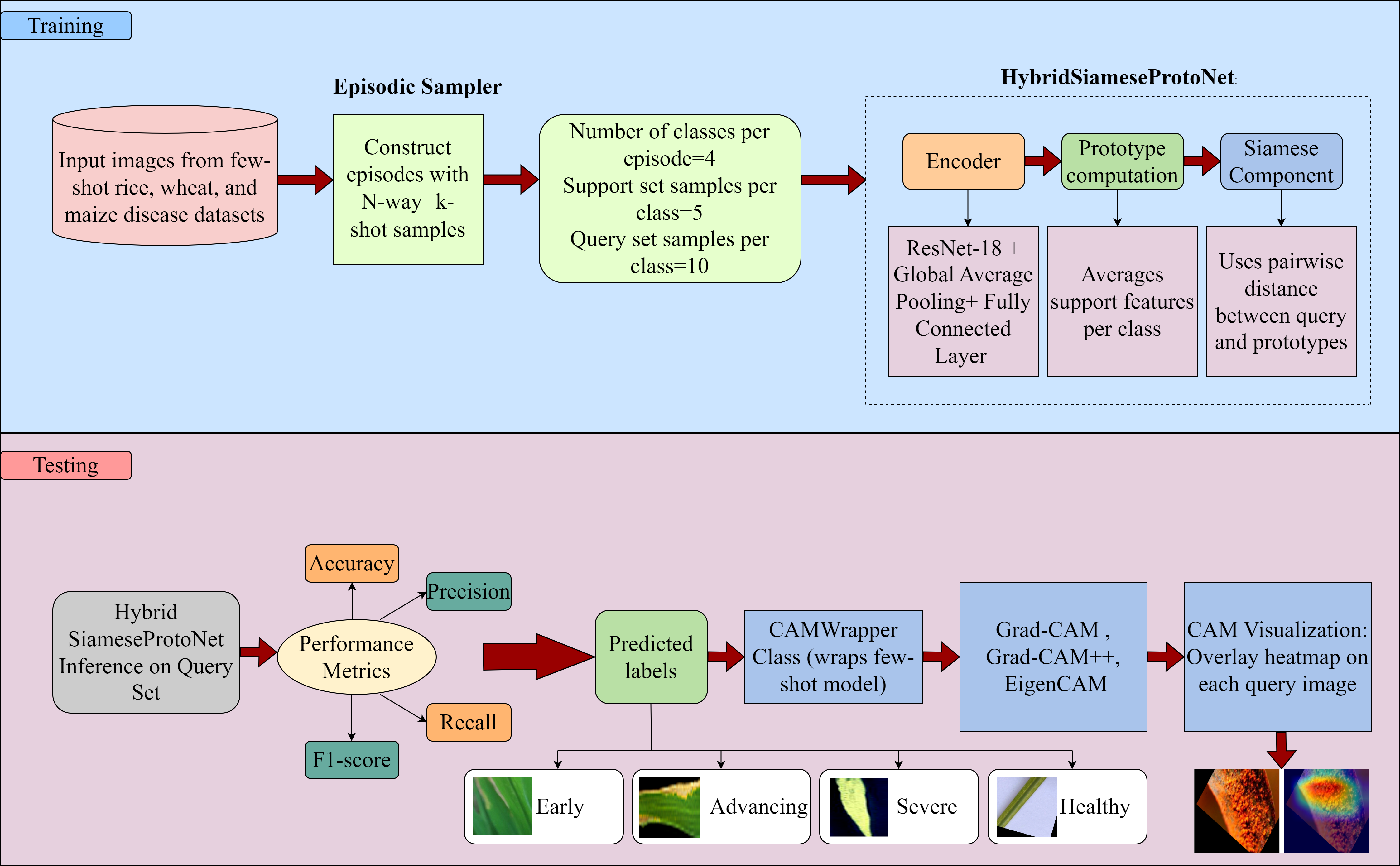}
}
\captionsetup{justification=centering}
\caption{The architecture of the Proposed Hybrid Siamese-ProtoNet Model}
\label{fig Hybrid Model}       % Give a unique label
\end{figure*}

\begin{table}[!t]
\centering
\caption{The Hyperparameters used for Hybrid Siamese-ProtoNet}
\label{tab:Hyperparameters}
\resizebox{8cm}{!}{%
\begin{tabular}{@{}lll@{}}
\toprule
\multicolumn{1}{c}{Parameter} & Value            & Decription                      \\ \midrule
num\_classes                   & 4                & Classes per episode             \\
num\_support                   & 5                & Support samples per class       \\
num\_query                     & 10               & Query samples per class         \\
feature\_dim                   & 512              & Output dimension of the encoder \\
optimizer                     & Adam             & Optimizer for training          \\
learning\_rate                 & le-4             & Learning rate for Adam          \\
loss\_fn                       & CrossEntropyLoss & Standard classification loss    \\
patience                      & 10               & Early stopping patience         \\
max\_epochs                    & 100              & Maximum training epochs         \\ \bottomrule
\end{tabular}%
}
\end{table}
%\end{itemize}
The architecture of the proposed Hybrid Siamese-ProtoNet model is illustrated in Fig.~\ref{fig Hybrid Model}. As shown in the figure, the few-shot datasets for rice, wheat, and maize are initially partitioned into training and validation sets. Episodic sampling is then employed to construct multiple episodes, each comprising four classes per episode. For each class, five samples are allocated to the support set and ten samples to the query set. This episodic framework enables the model to effectively learn class representations and generalize to new disease categories with limited data. \\

In the proposed few-shot learning framework, the ResNet-18 is used as the encoder, the last two layers of the model are removed, and customized average pooling and fully connected layers are added. The encoder uses the convolution layers for spatial feature maps, the global average pooling layers compress spatial features to a vector, and the fully connected layer is used to reduce dimensionality

During the forward pass of the model, the support and the query images are encoded, and prototypes are generated by computing the Mean of features from all support images of the same class. Euclidean distance between each query and all class prototypes is calculated and later converted to similarity. The performance of the model after training is evaluated on the test set, with the performance metrics accuracy, precision, recall, and F1-score. The CAMWrapper is used for interoperability; it allows Grad-CAM to interact with a multi-input model. In the visualization phase, it generates a few-shot episode from test data, wraps the model with CAMWrapper, applies Grad-CAM, GradCAM++, and EigenCAM to each query image, and Overlays the CAM heatmaps on the query image for interoperability. The overall framework of the proposed hybrid few-shot based model combines similarity-based comparison, classification by measuring distance to class prototypes, and interpretable visualization of model attention.
\section{Experimental Results}

\subsection{Siamese Network} 

The few-shot datasets formed for the rice, wheat, and maize crops were successfully trained and tested on the Siamese model proposed in the study. The model was able to efficiently learn the feature embeddings of most of the samples of the datasets developed. The results of the proposed model on the few-shot dataset for the disease stages of the rice plant are summarized in Table~\ref{s1}.

\begin{table*}[!h]
	\centering
	\caption{Performance of the Siamese Model on the Disease Stages of Few-shot Plant Leaf Datasets (5 Samples per Class)}
	\label{s1}
	\resizebox{16cm}{!}{%
		\begin{tabular}{@{}llllllllll@{}}
			\toprule
			Plant Type &
			Disease &
			\begin{tabular}[c]{@{}l@{}}No of \\ Classes\end{tabular} &
			\begin{tabular}[c]{@{}l@{}}Support Set \\ Samples\end{tabular} &
			\begin{tabular}[c]{@{}l@{}}Query Set \\ Samples\end{tabular} &
			TA &
			VA &
			Precision &
			Recall &
			F1-Score \\ \midrule
			
			\multirow{4}{*}{Rice} 
			& Bacterial Blight       & 4 & 5 & 5 & 0.8500 & 0.8000 & 1.0000 & 0.8000 & 0.8889 \\
			& Bacterial Leaf Streak  & 4 & 5 & 5 & 0.9000 & 0.8500 & 1.0000 & 0.8500 & 0.9189 \\
			& Blast                  & 4 & 5 & 5 & 0.9200 & 0.9000 & 1.0000 & 0.9000 & 0.9474 \\
			& Brown Spot             & 4 & 5 & 5 & 0.7500 & 0.7000 & 1.0000 & 0.7000 & 0.8235 \\ \midrule
			
			\multirow{4}{*}{Wheat} 
			& Leaf Rust              & 4 & 5 & 5 & 0.9500 & 0.9000 & 1.0000 & 0.9000 & 0.9474 \\
			& Powdery Mildew         & 4 & 5 & 5 & 0.7000 & 0.6000 & 1.0000 & 0.6000 & 0.7500 \\
			& Stripe Rust            & 4 & 5 & 5 & 0.7800 & 0.6500 & 1.0000 & 0.6500 & 0.7879 \\
			& Tan Spot               & 4 & 5 & 5 & 0.7500 & 0.6500 & 1.0000 & 0.6500 & 0.7879 \\ \midrule
			
			\multirow{4}{*}{Maize} 
			& Common Rust            & 4 & 5 & 5 & 0.9500 & 0.9000 & 1.0000 & 0.9000 & 0.9474 \\
			& Grey Leaf Spot         & 4 & 5 & 5 & 0.8000 & 0.7000 & 1.0000 & 0.7000 & 0.8235 \\
			& Northern Leaf Blight   & 4 & 5 & 5 & 0.9000 & 0.8500 & 1.0000 & 0.8500 & 0.9189 \\
			& Southern Rust          & 4 & 5 & 5 & 0.9000 & 0.8000 & 1.0000 & 0.8000 & 0.8889 \\ 
			
			\bottomrule
		\end{tabular}%
	}
\end{table*}

\begin{table*}[!h]
	\centering
	\caption{Performance of the Siamese Model on the Disease Stages of Few-shot Plant Leaf Datasets (80 Samples per Class)}
	\label{S80}
	\resizebox{16cm}{!}{%
		\begin{tabular}{@{}llllllllll@{}}
			\toprule
			Plant Type &
			Disease &
			\begin{tabular}[c]{@{}l@{}}No of \\ Classes\end{tabular} &
			\begin{tabular}[c]{@{}l@{}}Support Set \\ Samples\end{tabular} &
			\begin{tabular}[c]{@{}l@{}}Query Set \\ Samples\end{tabular} &
			TA &
			VA &
			Precision &
			Recall &
			F1-Score \\ \midrule
			
			\multirow{4}{*}{Rice} 
			& Bacterial Blight       & 4 & 80 & 10 & 0.9800 & 0.9500 & 1.0000 & 0.9500 & 0.9740 \\
			& Bacterial Leaf Streak  & 4 & 80 & 10 & 0.9900 & 0.9870 & 1.0000 & 0.9870 & 0.9576 \\
			& Blast                  & 4 & 80 & 10 & 0.9000 & 0.9750 & 1.0000 & 0.9750 & 0.9873 \\
			& Brown Spot             & 4 & 80 & 10 & 0.9500 & 0.9000 & 1.0000 & 0.9000 & 0.9474 \\ \midrule
			
			\multirow{4}{*}{Wheat} 
			& Leaf Rust              & 4 & 80 & 10 & 0.9800 & 0.9000 & 1.0000 & 0.9000 & 0.9740 \\
			& Powdery Mildew         & 4 & 80 & 10 & 0.8500 & 0.7000 & 1.0000 & 0.7000 & 0.8235 \\
			& Stripe Rust            & 4 & 80 & 10 & 0.8500 & 0.7500 & 1.0000 & 0.7500 & 0.8571 \\
			& Tan Spot               & 4 & 80 & 10 & 0.9500 & 0.8500 & 1.0000 & 0.8500 & 0.9189 \\ \midrule
			
			\multirow{4}{*}{Maize} 
			& Common Rust            & 4 & 80 & 10 & 0.9500 & 0.9870 & 1.0000 & 0.9870 & 0.9576 \\
			& Grey Leaf Spot         & 4 & 80 & 10 & 0.9000 & 0.8250 & 1.0000 & 0.8250 & 0.9041 \\
			& Northern Leaf Blight   & 4 & 80 & 10 & 0.9800 & 0.8750 & 1.0000 & 0.8750 & 0.9333 \\
			& Southern Rust          & 4 & 80 & 10 & 0.9000 & 0.8000 & 1.0000 & 0.8000 & 0.8889 \\ 
			
			\bottomrule
		\end{tabular}%
	}
\end{table*}

\begin{table*}[!h]
	\centering
	\caption{Performance of the Relational Model on the Disease Stages of the Few-shot Plant Leaf Datasets (5 Samples per Class)}
	\label{R1}
	\resizebox{16cm}{!}{%
		\begin{tabular}{@{}lllllllll@{}}
			\toprule
			Plant Type &
			Disease &
			\begin{tabular}[c]{@{}l@{}}No of \\ Classes\end{tabular} &
			\begin{tabular}[c]{@{}l@{}}Support Set \\ Samples\end{tabular} &
			\begin{tabular}[c]{@{}l@{}}Query Set \\ Samples\end{tabular} &
			VA &
			Precision &
			Recall &
			F1-Score \\ \midrule
			
			\multirow{4}{*}{Rice} 
			& Bacterial Blight       & 4 & 5 & 5 & 0.9733 & 0.9734 & 0.9733 & 0.9733 \\
			& Bacterial Leaf Streak  & 4 & 5 & 5 & 0.9780 & 0.9772 & 0.9780 & 0.9780 \\
			& Blast                  & 4 & 5 & 5 & 0.9784 & 0.9786 & 0.9784 & 0.9784 \\
			& Brown Spot             & 4 & 5 & 5 & 0.8500 & 0.8499 & 0.8500 & 0.8500 \\ \midrule
			
			\multirow{4}{*}{Wheat} 
			& Leaf Rust              & 4 & 5 & 5 & 0.9825 & 0.9826 & 0.9825 & 0.9825 \\
			& Powdery Mildew         & 4 & 5 & 5 & 0.9500 & 0.9572 & 0.9500 & 0.9495 \\
			& Stripe Rust            & 4 & 5 & 5 & 0.9880 & 0.9882 & 0.9880 & 0.9880 \\
			& Tan Spot               & 4 & 5 & 5 & 0.7930 & 0.7938 & 0.7930 & 0.7933 \\ \midrule
			
			\multirow{4}{*}{Maize} 
			& Common Rust            & 4 & 5 & 5 & 0.9890 & 0.9891 & 0.9890 & 0.9890 \\
			& Grey Leaf Spot         & 4 & 5 & 5 & 0.9680 & 0.9682 & 0.9680 & 0.9680 \\
			& Northern Leaf Blight   & 4 & 5 & 5 & 0.9979 & 0.9981 & 0.9979 & 0.9979 \\
			& Southern Rust          & 4 & 5 & 5 & 0.9776 & 0.9778 & 0.9776 & 0.9776 \\ 
			
			\bottomrule
		\end{tabular}%
	}
\end{table*}

\begin{table*}[!h]
	\centering
	\caption{Performance of the Matching Networks on the Disease Stages of the Few-shot Plant Leaf Datasets (30 Samples per Class)}
	\label{M1}
	\resizebox{16cm}{!}{%
		\begin{tabular}{@{}lllllllll@{}}
			\toprule
			Plant Type &
			Disease &
			\begin{tabular}[c]{@{}l@{}}No of \\ Classes\end{tabular} &
			\begin{tabular}[c]{@{}l@{}}Support Set \\ Samples\end{tabular} &
			\begin{tabular}[c]{@{}l@{}}Query Set \\ Samples\end{tabular} &
			\begin{tabular}[c]{@{}l@{}}Validation\\ Accuracy\end{tabular} &
			Precision &
			Recall &
			F1-Score \\ \midrule
			
			\multirow{4}{*}{Rice} 
			& Bacterial Blight       & 4 & 30 & 30 & 0.6380 & 0.6382 & 0.6380 & 0.6380 \\
			& Bacterial Leaf Streak  & 4 & 30 & 30 & 0.7230 & 0.7232 & 0.7230 & 0.7230 \\
			& Blast                  & 4 & 30 & 30 & 0.6669 & 0.6671 & 0.6669 & 0.6669 \\
			& Brown Spot             & 4 & 30 & 30 & 0.7932 & 0.7934 & 0.7932 & 0.7932 \\ \midrule
			
			\multirow{4}{*}{Wheat} 
			& Leaf Rust              & 4 & 30 & 30 & 0.6118 & 0.5915 & 0.6118 & 0.5963 \\
			& Powdery Mildew         & 4 & 30 & 30 & 0.7217 & 0.7197 & 0.7217 & 0.7062 \\
			& Stripe Rust            & 4 & 30 & 30 & 0.6449 & 0.6257 & 0.6449 & 0.6202 \\
			& Tan Spot               & 4 & 30 & 30 & 0.6662 & 0.6615 & 0.6662 & 0.6453 \\ \midrule
			
			\multirow{4}{*}{Maize} 
			& Common Rust            & 4 & 30 & 30 & 0.7790 & 0.7791 & 0.7790 & 0.7790 \\
			& Grey Leaf Spot         & 4 & 30 & 30 & 0.6640 & 0.6642 & 0.6640 & 0.6640 \\
			& Northern Leaf Blight   & 4 & 30 & 30 & 0.7369 & 0.7371 & 0.7369 & 0.7369 \\
			& Southern Rust          & 4 & 30 & 30 & 0.6543 & 0.6545 & 0.6543 & 0.6543 \\ 
			
			\bottomrule
		\end{tabular}%
	}
\end{table*}

\begin{table*}[!h]
	\centering
	\caption{Performance of the Prototypical Model on the Disease Stages of the Few-shot Plant Leaf Datasets (5 Support, 10 Query per Class)}
	\label{P}
	\resizebox{16cm}{!}{%
		\begin{tabular}{@{}lllllllll@{}}
			\toprule
			Plant Type &
			Disease &
			\begin{tabular}[c]{@{}l@{}}No of \\ Classes\end{tabular} &
			\begin{tabular}[c]{@{}l@{}}Support Set \\ Samples\end{tabular} &
			\begin{tabular}[c]{@{}l@{}}Query Set \\ Samples\end{tabular} &
			\begin{tabular}[c]{@{}l@{}}Validation\\ Accuracy\end{tabular} &
			Precision &
			Recall &
			F1-Score \\ \midrule
			
			\multirow{4}{*}{Rice}
			& Bacterial Blight       & 4 & 5 & 10 & 0.9746 & 0.9748 & 0.9746 & 0.9746 \\
			& Bacterial Leaf Streak  & 4 & 5 & 10 & 0.9512 & 0.9514 & 0.9512 & 0.9512 \\
			& Blast                  & 4 & 5 & 10 & 0.9747 & 0.9749 & 0.9747 & 0.9747 \\
			& Brown Spot             & 4 & 5 & 10 & 0.9688 & 0.9690 & 0.9688 & 0.9688 \\ \midrule
			
			\multirow{4}{*}{Wheat}
			& Leaf Rust              & 4 & 5 & 10 & 0.9536 & 0.9538 & 0.9536 & 0.9536 \\
			& Powdery Mildew         & 4 & 5 & 10 & 0.9400 & 0.9402 & 0.9400 & 0.9400 \\
			& Stripe Rust            & 4 & 5 & 10 & 0.9549 & 0.9552 & 0.9549 & 0.9549 \\
			& Tan Spot               & 4 & 5 & 10 & 0.9612 & 0.9614 & 0.9612 & 0.9612 \\ \midrule
			
			\multirow{4}{*}{Maize}
			& Common Rust            & 4 & 5 & 10 & 0.9876 & 0.9878 & 0.9876 & 0.9876 \\
			& Grey Leaf Spot         & 4 & 5 & 10 & 0.9663 & 0.9665 & 0.9663 & 0.9663 \\
			& Northern Leaf Blight   & 4 & 5 & 10 & 0.9577 & 0.9579 & 0.9577 & 0.9577 \\
			& Southern Rust          & 4 & 5 & 10 & 0.9886 & 0.9888 & 0.9886 & 0.9886 \\ 
			
			\bottomrule
		\end{tabular}%
	}
\end{table*}

\begin{table*}[!h]
	\centering
	\caption{Performance of the Hybrid Siamese-ProtoNet Model on the Disease Stages of the Few-shot Plant Leaf Datasets (5 Support, 10 Query per Class)}
	\label{tab:hybrid-merged}
	\resizebox{16cm}{!}{%
		\begin{tabular}{@{}lllllllll@{}}
			\toprule
			Plant Type &
			Disease &
			\begin{tabular}[c]{@{}l@{}}No of \\ Classes\end{tabular} &
			\begin{tabular}[c]{@{}l@{}}Support Set \\ Samples\end{tabular} &
			\begin{tabular}[c]{@{}l@{}}Query Set \\ Samples\end{tabular} &
			\begin{tabular}[c]{@{}l@{}}Validation\\ Accuracy\end{tabular} &
			Precision &
			Recall &
			F1-Score \\ \midrule
			
			\multirow{4}{*}{Rice}
			& Bacterial Blight       & 4 & 5 & 10 & 0.9750 & 0.9773 & 0.9750 & 0.9749 \\
			& Bacterial Leaf Streak  & 4 & 5 & 10 & 0.9750 & 0.9773 & 0.9750 & 0.9749 \\
			& Blast                  & 4 & 5 & 10 & 0.9750 & 0.9773 & 0.9750 & 0.9749 \\
			& Brown Spot             & 4 & 5 & 10 & 0.9750 & 0.9773 & 0.9750 & 0.9749 \\ \midrule
			
			\multirow{4}{*}{Wheat}
			& Leaf Rust              & 4 & 5 & 10 & 0.9550 & 0.9568 & 0.9550 & 0.9548 \\
			& Powdery Mildew         & 4 & 5 & 10 & 0.9550 & 0.9568 & 0.9550 & 0.9548 \\
			& Stripe Rust            & 4 & 5 & 10 & 0.9750 & 0.9773 & 0.9750 & 0.9749 \\
			& Tan Spot               & 4 & 5 & 10 & 0.9750 & 0.9773 & 0.9750 & 0.9749 \\ \midrule
			
			\multirow{4}{*}{Maize}
			& Common Rust            & 4 & 5 & 10 & 0.9250 & 0.9245 & 0.9250 & 0.9236 \\
			& Grey Leaf Spot         & 4 & 5 & 10 & 0.9750 & 0.9773 & 0.9750 & 0.9749 \\
			& Northern Leaf Blight   & 4 & 5 & 10 & 0.9750 & 0.9773 & 0.9750 & 0.9749 \\
			& Southern Rust          & 4 & 5 & 10 & 0.9250 & 0.9245 & 0.9250 & 0.9236 \\ 
			
			\bottomrule
		\end{tabular}%
	}
\end{table*}

\begin{figure*}[!h]
\centering
\subfloat[]{\includegraphics[width=1.5in]{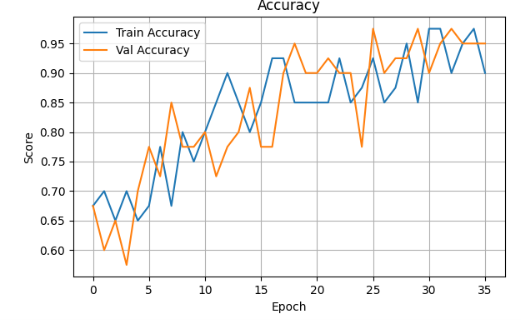}%
\label{Accuracy}}
\hfil
\subfloat[]{\includegraphics[width=1.5in]{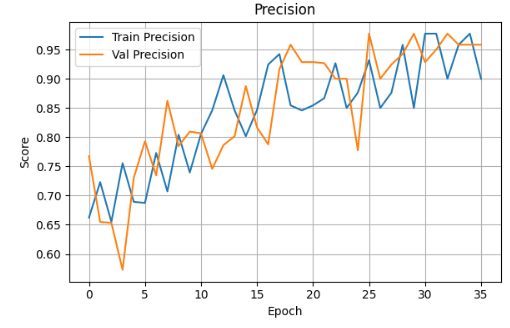}%
\label{Precision}}
\hfil
\subfloat[]{\includegraphics[width=1.5in]{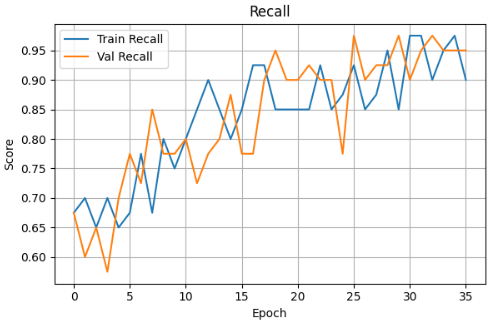}%
\label{Recall}}
\hfil
\subfloat[]{\includegraphics[width=1.5in]{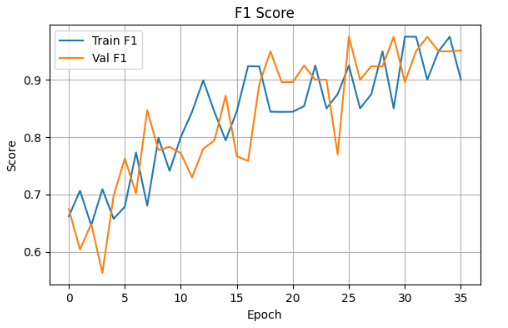}%
\label{F1-score}}
\caption{The Accuracy, Precision, Recall, and F1-score of the Hybrid Siamese-ProtoNet Model on the Disease Stages of Rice Bacterial Blight Disease. (a) Accuracy, (b) Precision, (c) Recall, (d) F1-score}
\label{fig 2}
\end{figure*}
\begin{figure*}[!h]
\centering
\subfloat[]{\includegraphics[width=1.5in]{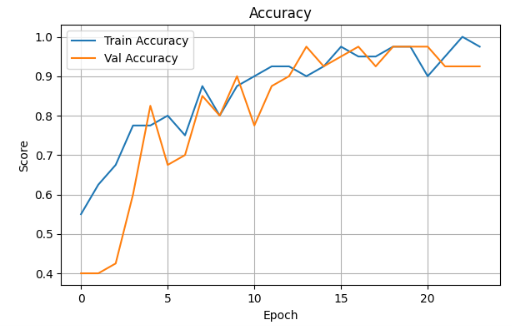}%
\label{Accuracy}}
\hfil
\subfloat[]{\includegraphics[width=1.5in]{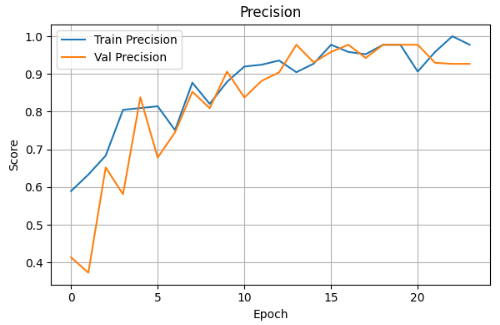}%
\label{Precision}}
\hfil
\subfloat[]{\includegraphics[width=1.5in]{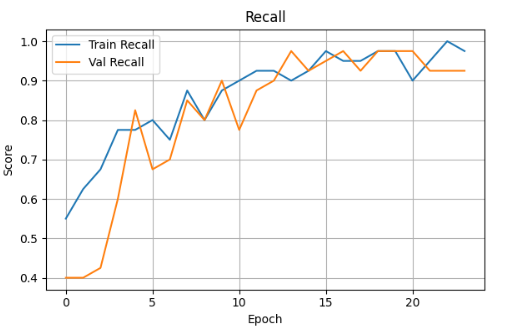}%
\label{Recall}}
\hfil
\subfloat[]{\includegraphics[width=1.5in]{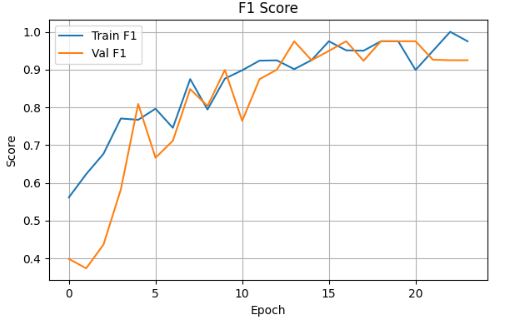}%
\label{F1-score}}
\caption{The Accuracy, Precision, Recall, and F1-score of the Hybrid Siamese-ProtoNet Model on the Disease Stages of Stripe Rust Disease of Wheat. (a) Accuracy, (b) Precision, (c) Recall, (d) F1-score}
\label{fig 3}
\end{figure*}
\begin{figure*}[!h]
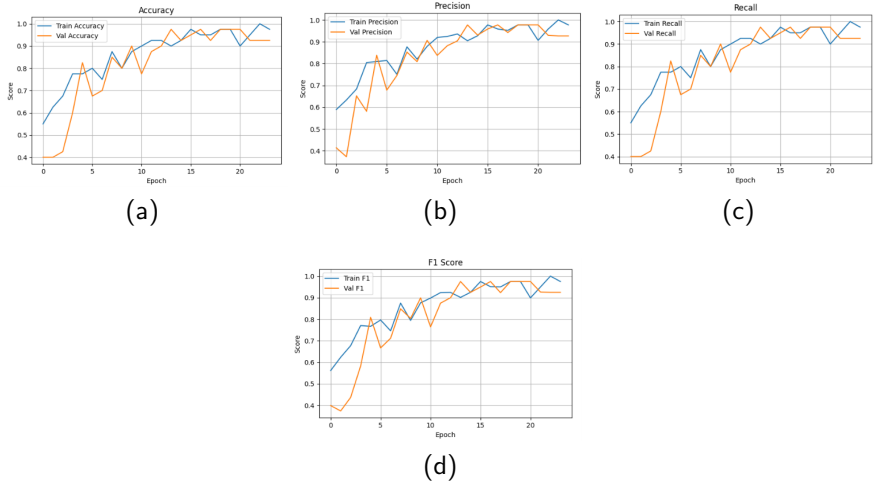

\centering
\subfloat[]{\includegraphics[width=1.5in]{ma.png}%
\label{Accuracy}}
\hfil
\subfloat[]{\includegraphics[width=1.5in]{mp.png}%
\label{Precision}}
\hfil
\subfloat[]{\includegraphics[width=1.5in]{mr.png}%
\label{Recall}}
\hfil
\subfloat[]{\includegraphics[width=1.5in]{mf1.png}
\label{F1-score}}
\caption{The Accuracy, Precision, Recall, and F1-score of the Hybrid Siamese-ProtoNet Model on the Disease Stages of Northern Leaf Blight Disease of Maize (a) Accuracy, (b) Precision, (c) Recall, (d) F1-score}
\label{fig 4}
\end{figure*}
\begin{figure*}[!h]
\center
\resizebox{10cm}{!}{% 
\includegraphics[width=\textwidth]{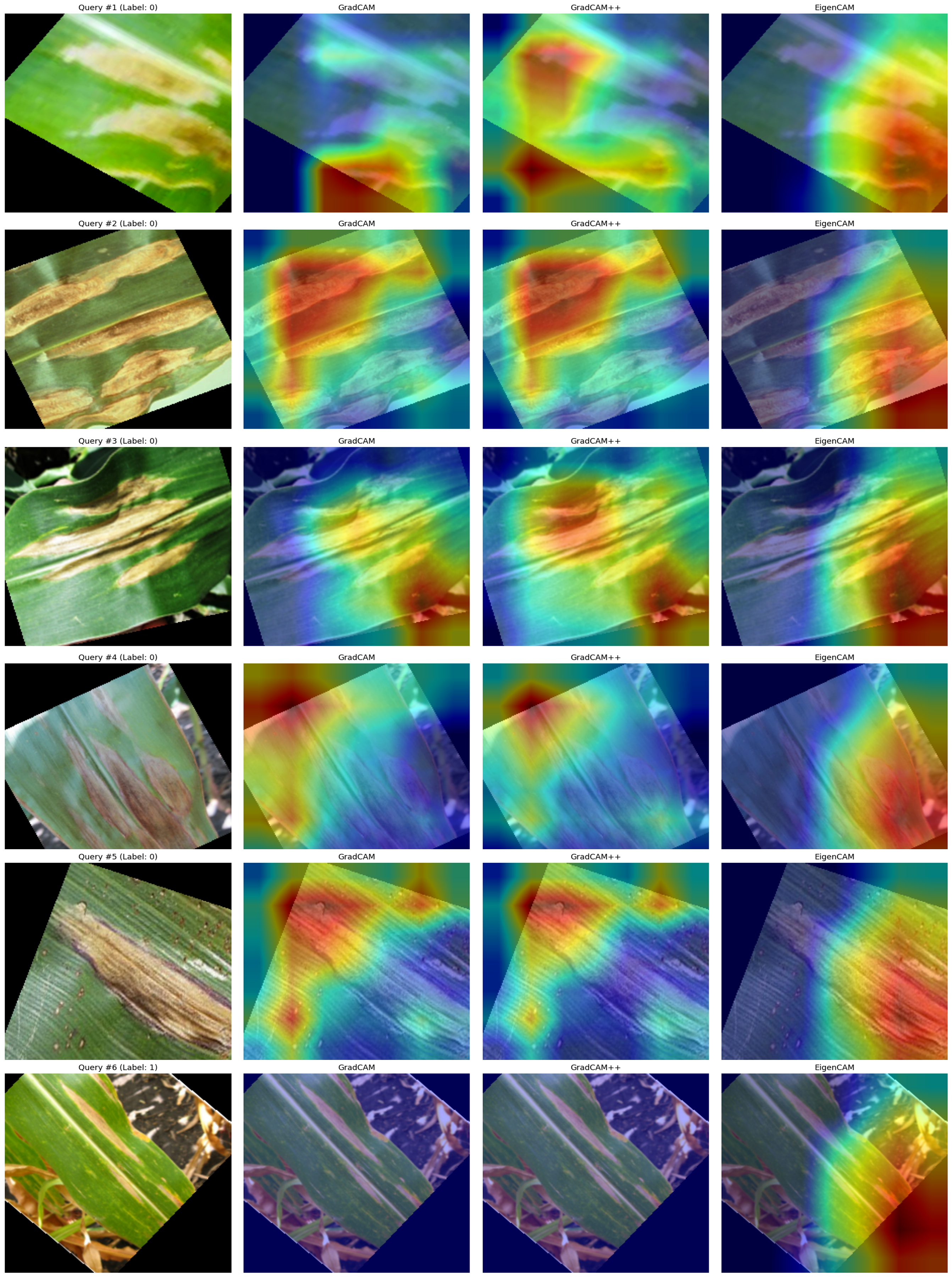}
}
\captionsetup{justification=centering}
\caption{CAM Visualization of the Hybrid Siamese-ProtoNet Model on the Disease Stages of the Northern Blight Disease of the Maize Plant \cite{robertson2014nclb, isakeit2021_nclb, jackson2012maize, malvick_nclb_image, ocj2014corn}}
\label{fig 5}       % Give a unique label
\end{figure*}
\begin{figure*}[!h]
\center
\resizebox{10cm}{!}{% 
\includegraphics[width=\textwidth]{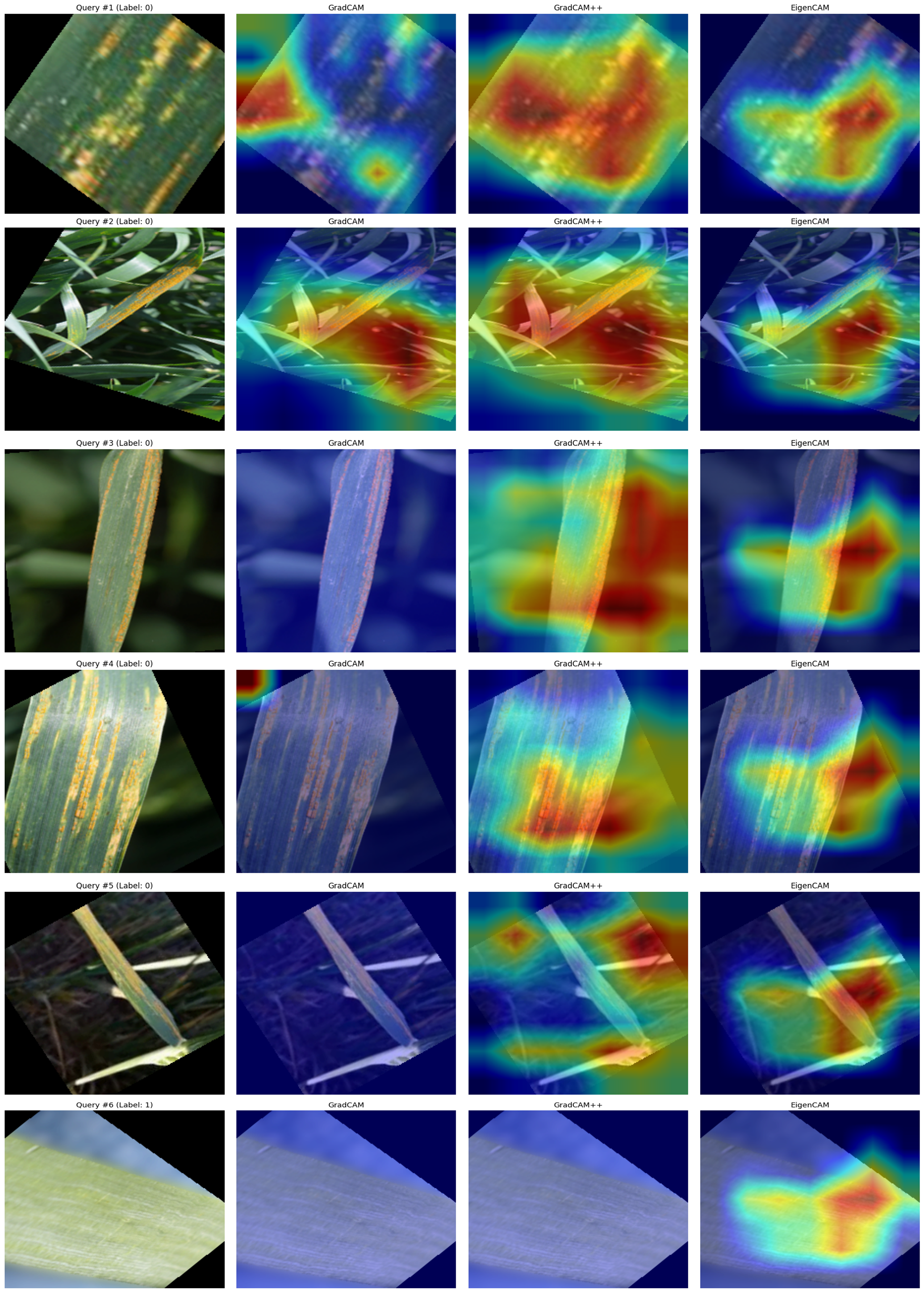}
}
\captionsetup{justification=centering}
\caption{CAM Visualization of the Hybrid Siamese-ProtoNet Model on the Disease Stages of the StripeRust Disease of the Wheat Plant}
\label{fig 6}       % Give a unique label
\end{figure*}
\begin{figure*}[!h]
\center
\resizebox{10cm}{!}{% 
\includegraphics[width=\textwidth]{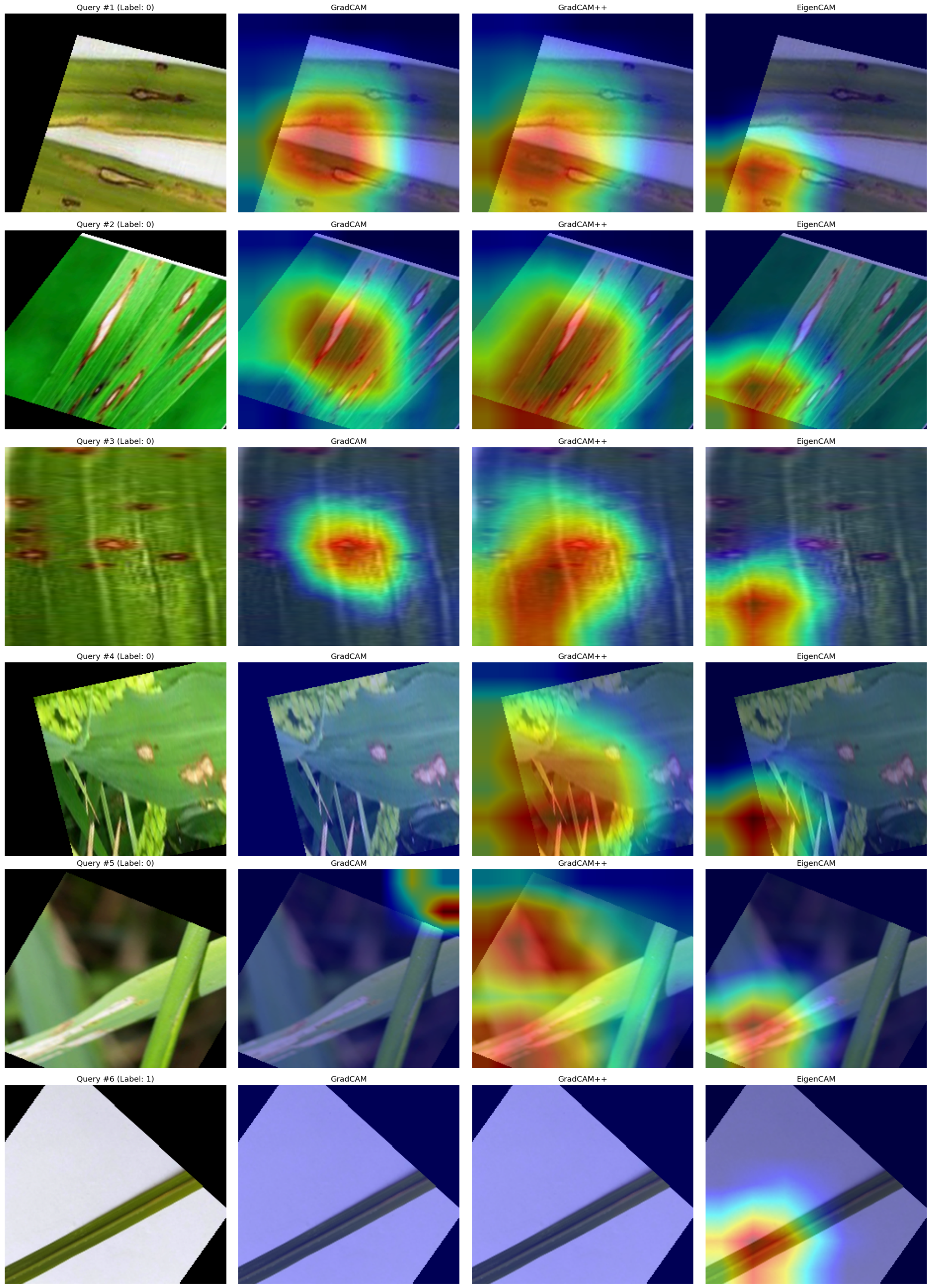}
}
\captionsetup{justification=centering}
\caption{CAM Visualization of the Hybrid Siamese-ProtoNet Model on the Disease Stages of the RiceBlast Disease of the Rice Plant}
\label{fig 7}       % Give a unique label
\end{figure*}
The performance of the Siamese model in the few-shot dataset formed to identify the disease stages of the wheat and maize plants is also shown in Table ~\ref{s1}.
%The performance of the Siamese model on the few-shot dataset formed to identify the disease stages of the maize plants are as shown in Table ~\ref{tab:3}
Table ~\ref{s1} shows the experimental results of the siamese model when it was trained with only five samples in the support set and five samples in the query set. Several experimentations were conducted to improve the performance of the model with 20, 30, 50, and 80 samples in the support set and 10 samples in the query set. The model performed best with 80 samples in the support set and 10 samples in the query set. Table ~\ref{S80} shows the performance of the siamese model when trained with 80 samples in the support set and 10 samples in the query set. The few-shot datasets developed for the three crop types, when evaluated using the proposed Siamese model, yielded accuracy, precision, recall, and F1-scores exceeding 70\% for rice, maize, and wheat. Among the maize diseases, the stages of Common Rust achieved the highest performance, with all evaluation metrics surpassing 95\%. In the case of rice, the stages of Bacterial Leaf Streak attained the highest validation accuracy, precision, recall, and F1-score, each exceeding 95\%. For wheat, the stages of Leaf Rust recorded the best performance, with all metrics above 90\%. The results show that the performance have improved when compared to the results in Table ~\ref{s1}.

\subsection{Relational Network} The few-shot datasets formed for rice, wheat, and maize crops were also tested on the proposed relational network. The results of the performance of the relational model on the few-shot wheat, rice, and rice datasets are shown in Table ~\ref{R1}. From the table, it can be observed that the relational model achieved validation accuracy, precision, recall, and F1-scores above 79\% across all disease stages in the few-shot wheat leaf dataset, with the highest performance—exceeding 98\%—observed for the Stripe Rust disease stages. For the few-shot rice leaf dataset, the model attained accuracy, precision, recall, and F1-scores above 84\%, with the Rice Blast disease stage demonstrating the best performance, surpassing 97\% across all metrics. In the case of maize, the model achieved over 96\% in all evaluation metrics, with the Northern Leaf Blight disease stages exhibiting the highest performance, exceeding 99\% in accuracy, precision, recall, and F1-score. Overall, the proposed relational model performed well on the few-shot datasets developed, similar to the Prototypical model and Siamese model developed in the work.

\subsection{Matching Networks} The network model designed for few-shot classification performed best when trained with thirty samples per class for the support set and thirty samples per class for the query set. The performance of the model failed to improve when experimented with only 5, 10, 15, and 20 samples per class in the support and query set; the accuracy of the model did not improve from 25\%. The pre-trained ResNet-18 model acted as a strong feature extractor. The model was evaluated on all three datasets developed. The performance of the matching networks on the datasets is as shown in Table ~\ref{M1}. Table ~\ref{M1} show the model's performance on the disease stages of the maize, rice, and wheat crops. From the results, it can be observed that the Maching Networks achieved validation accuracy, precision, recall, and F1-scores above 64\% across all disease stages in the few-shot wheat leaf dataset, with the highest performance—exceeding 70\%—observed for the Powdery Mildew disease stages. For the few-shot rice leaf dataset, the model attained accuracy, precision, recall, and F1-scores above 63\%, with the Rice Brown Spot disease stage demonstrating the best performance, surpassing 79\% across all metrics. In the case of maize, the model achieved over 65\% in all evaluation metrics, with the Common Rust disease stages exhibiting the highest performance, exceeding 77\% in accuracy, precision, recall, and F1-score.  Which is comparatively less when compared to the performance of the proposed Siamese, Relational, and Prototypical models on the few-shot datasets of maize, rice, and wheat crops.

\subsection{Prototypical Networks} The performance of the proposed model for few-shot image classifications for disease stages of the diseases considered in the research for wheat, rice, and maize plants is shown in Table~\ref{P}. The model was able to obtain an accuracy of around 94\% for the wheat leaf dataset, above 95\% for the rice leaf dataset, and above 95\% for the maize leaf dataset with episodic training and only a few number of training samples, the pre-trained ResNet 18 model used as the backbone was efficient in extracting the features from the samples. Therefore, the model performed fairly well on all three few-shot datasets considered in the research. The prototypical network developed in this study achieved validation accuracy, precision, recall, and F1-scores exceeding 94\% on the leaf datasets, with the highest performance observed for the disease stages of Tan Spot. For the few-shot rice leaf dataset, the model attained accuracy, precision, recall, and F1-scores above 95\% across all considered diseases, with Rice Blast exhibiting the best results, surpassing 97\% in all performance metrics. Similarly, on the maize leaf dataset, the prototypical network consistently achieved over 95\% across all evaluation metrics for every disease class. Notably, the disease stages of Southern Rust demonstrated the highest performance, with validation accuracy, precision, recall, and F1-scores all exceeding 98\%. 

\subsection{Proposed Hybrid Siamese-ProtoNet Model} The hybrid model developed in this study was evaluated on the few-shot datasets constructed for the disease stages of rice, wheat, and maize crops, in a manner consistent with other few-shot learning approaches. The performance of the proposed model on the disease stages of the few-shot rice leaf dataset is shown in Table~\ref{tab:hybrid-merged}

The results show that the model was able to obtain an accuracy, precision, recall, and F1-score of above 97\% for all disease stages of the few-shot rice leaf dataset. The model demonstrated superior performance across all rice plant diseases compared to the results achieved by the individual Siamese network, Prototypical Network, Relation Network, and Matching Networks. The proposed model achieved a significantly lower execution time of approximately two hours, outperforming the other models, which required longer processing times. The accuracy, precision, recall, and F1-score of the proposed hybrid model on the disease stages of the Rice Bacterial Blight disease of rice plants are as shown in Fig.~\ref{fig 2}
An accuracy, precision, and recall of above 95\% was obtained for all the disease stages of the diseases considered for the wheat plants in the study. The  Table ~\ref{tab:hybrid-merged} also represents the performance of the proposed hybrid Siamese-Protonet model on the few-shot wheat leaf dataset. 
The results demonstrate that the proposed model achieved consistently strong performance across all wheat disease stages when compared to existing few-shot learning approaches, including Matching Networks, Prototypical Networks, the Siamese Network, and the Relation Network. These models were evaluated using the few-shot datasets specifically developed in this study. The accuracy, precision, recall, and F1-score obtained by the proposed model on the Stripe Rust Disease Stages of wheat are presented in Fig.~\ref{fig 3}, highlighting its effectiveness in addressing the classification task.
An accuracy, precision, and recall of above 92\% was obtained for all the disease stages of the diseases considered for the maize plants in the study. The  Table ~\ref{tab:hybrid-merged} represents the performance of the proposed hybrid Siamese-Protonet model on the few-shot maize leaf dataset also. The model performed well when compared the matching networks, and the Siamese model. The model was able to obtain a performance metric similar to that obtained by the Prototypical Network, and Relational model used in the study. The accuracy, precision, recall, and F1-score obtained by the proposed model on the Northern Leaf Blight Disease  stages of maize are presented in Fig.~\ref{fig 4}.
CAM visualaizations on the Northern Leaf Blight disease stages of the maize plant is as shown in Fig.~\ref{fig 5}. CAM visualaizations  on the StripeRust disease stages of the wheat plant is as shown in Fig.~\ref{fig 6}. CAM visualaizations  on RiceBlast disease stages of the rice plant is as shown in Fig.~\ref{fig 7}.

\section{Conclusion}
In this paper, we have introduced a hybrid Siamese-ProtoNet model. The proposed model demonstrated above-average performance across all disease stages for rice, wheat, and maize, even when trained and tested on very limited samples. Its performance is comparable to that of several deep learning models reported in existing literature. To enhance interpretability and gain insights into the model’s decision-making process, Grad-CAM is employed, effectively highlighting the key regions influencing predictions. The custom few-shot datasets developed in this study were also evaluated using baseline few-shot learning models, including Siamese, Prototypical, Relation, and Matching Networks, all of which performed reasonably well. The proposed hybrid model outperformed or matched their performance, showcasing its robustness under data-constrained conditions. In future work, the dataset can be extended by incorporating additional crop diseases with limited annotated samples to further evaluate and refine the model’s generalizability and effectiveness.
\section{Acknowledgement}
The authors gratefully acknowledge that this research work was carried out in the Creative Lab, BITS Pilani Dubai Campus, Dubai, UAE. The support, resources, and research environment provided by the institution are sincerely appreciated and contributed significantly to the successful completion of this study.
\section{Declarations}
\begin{itemize}
	\item Availability of data and material: Not applicable
	\item Competing interests: Not applicable
	\item Funding: Not applicable
	\item Authors' contributions: Diana Susane Joseph directed the study's conception and design, conducted a comprehensive literature review, structured the taxonomy of challenges and strategies, and authored the initial draft of the manuscript. Pranav M. Pawar contributed to the drafting and enhancement of key sections of the manuscript, supervision, oversaw the research process, and participated in the analysis of results. Raja M offered essential feedback, assessed and refined the manuscript for technical rigor and clarity, and guided the research trajectory. Mithun Mukherjee engaged in manuscript revision, facilitated the critical assessment of existing methodologies, and contributed to discussions regarding unresolved issues and prospective research directions. 
	\item Acknowledgements: Not applicable
\end{itemize}
\bibliographystyle{splncs04}
\bibliography{ref}

\begin{thebibliography}{10}

\bibitem{arsenovic2019solving}
M.~Arsenovic, M.~Karanovic, S.~Sladojevic, A.~Anderla, and D.~Stefanovic,
  ``Solving current limitations of deep learning based approaches for plant
  disease detection,'' {\em Symmetry}, vol.~11, no.~7, p.~939, 2019.

\bibitem{joseph2023intelligent}
D.~S. Joseph, P.~M. Pawar, and R.~Pramanik, ``Intelligent plant disease
  diagnosis using convolutional neural network: a review,'' {\em Multimedia
  Tools and Applications}, vol.~82, no.~14, pp.~21415--21481, 2023.

\bibitem{joseph2024plant}
D.~S. Joseph and P.~M. Pawar, ``Plant disease identification using efficient
  fine-tuning of deep learning models,'' in {\em 2024 4th International
  Conference on Artificial Intelligence and Signal Processing (AISP)},
  pp.~1--5, IEEE, 2024.

\bibitem{nagaraju2020systematic}
M.~Nagaraju and P.~Chawla, ``Systematic review of deep learning techniques in
  plant disease detection,'' {\em International journal of system assurance
  engineering and management}, vol.~11, no.~3, pp.~547--560, 2020.

\bibitem{oad2024plant}
A.~Oad, S.~S. Abbas, A.~Zafar, B.~A. Akram, F.~Dong, M.~S.~H. Talpur, and
  M.~Uddin, ``Plant leaf disease detection using ensemble learning and
  explainable ai,'' {\em IEEE Access}, 2024.

\bibitem{mindhe2020plant}
O.~Mindhe, O.~Kurkute, S.~Naxikar, and N.~Raje, ``Plant disease detection using
  deep learning,'' {\em International Research Journal of Engineering and
  Technology}, pp.~2497--2503, 2020.

\bibitem{ramesh2018plant}
S.~Ramesh, R.~Hebbar, M.~Niveditha, R.~Pooja, N.~Shashank, P.~Vinod, {\em
  et~al.}, ``Plant disease detection using machine learning,'' in {\em 2018
  International conference on design innovations for 3Cs compute communicate
  control (ICDI3C)}, pp.~41--45, IEEE, 2018.

\bibitem{ahmed2023plant}
I.~Ahmed and P.~K. Yadav, ``Plant disease detection using machine learning
  approaches,'' {\em Expert Systems}, vol.~40, no.~5, p.~e13136, 2023.

\bibitem{ahmad2023survey}
A.~Ahmad, D.~Saraswat, and A.~El~Gamal, ``A survey on using deep learning
  techniques for plant disease diagnosis and recommendations for development of
  appropriate tools,'' {\em Smart Agricultural Technology}, vol.~3, p.~100083,
  2023.

\bibitem{ferentinos2018deep}
K.~P. Ferentinos, ``Deep learning models for plant disease detection and
  diagnosis,'' {\em Computers and electronics in agriculture}, vol.~145,
  pp.~311--318, 2018.

\bibitem{joseph2024mobile}
D.~S. Joseph and P.~M. Pawar, ``Mobile-xcep hybrid model for plant disease
  diagnosis,'' {\em Multimedia Tools and Applications}, pp.~1--44, 2024.

\bibitem{rezaei2024plant}
M.~Rezaei, D.~Diepeveen, H.~Laga, M.~G. Jones, and F.~Sohel, ``Plant disease
  recognition in a low data scenario using few-shot learning,'' {\em Computers
  and electronics in agriculture}, vol.~219, p.~108812, 2024.

\bibitem{sun2024few}
J.~Sun, W.~Cao, X.~Fu, S.~Ochi, and T.~Yamanaka, ``Few-shot learning for plant
  disease recognition: A review,'' {\em Agronomy Journal}, vol.~116, no.~3,
  pp.~1204--1216, 2024.

\bibitem{li2021semi}
Y.~Li and X.~Chao, ``Semi-supervised few-shot learning approach for plant
  diseases recognition,'' {\em Plant Methods}, vol.~17, pp.~1--10, 2021.

\bibitem{argueso2020few}
D.~Arg{\"u}eso, A.~Picon, U.~Irusta, A.~Medela, M.~G. San-Emeterio,
  A.~Bereciartua, and A.~Alvarez-Gila, ``Few-shot learning approach for plant
  disease classification using images taken in the field,'' {\em Computers and
  Electronics in Agriculture}, vol.~175, p.~105542, 2020.

\bibitem{lin2022few}
H.~Lin, R.~Tse, S.-K. Tang, Z.-p. Qiang, and G.~Pau, ``Few-shot learning
  approach with multi-scale feature fusion and attention for plant disease
  recognition,'' {\em Frontiers in Plant Science}, vol.~13, p.~907916, 2022.

\bibitem{uskaner2024efficient}
P.~Uskaner~Hepsa{\u{g}}, ``Efficient plant disease identification using
  few-shot learning: a transfer learning approach,'' {\em Multimedia Tools and
  Applications}, vol.~83, no.~20, pp.~58293--58308, 2024.

\bibitem{joseph2024rice}
D.~S. Joseph and P.~M. Pawar, ``Rice leaf disease stages classification using
  few-shot learning techniques,'' in {\em 2024 International Conference on
  Smart Systems for Electrical, Electronics, Communication and Computer
  Engineering (ICSSEECC)}, pp.~13--18, IEEE, 2024.

\bibitem{cap2020leafgan}
Q.~H. Cap, H.~Uga, S.~Kagiwada, and H.~Iyatomi, ``Leafgan: An effective data
  augmentation method for practical plant disease diagnosis,'' {\em IEEE
  Transactions on Automation Science and Engineering}, vol.~19, no.~2,
  pp.~1258--1267, 2020.

\bibitem{qadri2024advances}
S.~A.~A. Qadri, N.-F. Huang, T.~M. Wani, and S.~A. Bhat, ``Advances and
  challenges in computer vision for image-based plant disease detection: a
  comprehensive survey of machine and deep learning approaches,'' {\em IEEE
  Transactions on Automation Science and Engineering}, 2024.

\bibitem{panchal2023image}
A.~V. Panchal, S.~C. Patel, K.~Bagyalakshmi, P.~Kumar, I.~R. Khan, and M.~Soni,
  ``Image-based plant diseases detection using deep learning,'' {\em Materials
  Today: Proceedings}, vol.~80, pp.~3500--3506, 2023.

\bibitem{mohameth2020plant}
F.~Mohameth, C.~Bingcai, and K.~A. Sada, ``Plant disease detection with deep
  learning and feature extraction using plant village,'' {\em Journal of
  Computer and Communications}, vol.~8, no.~6, pp.~10--22, 2020.

\bibitem{chowdhury2021automatic}
M.~E. Chowdhury, T.~Rahman, A.~Khandakar, M.~A. Ayari, A.~U. Khan, M.~S. Khan,
  N.~Al-Emadi, M.~B.~I. Reaz, M.~T. Islam, and S.~H.~M. Ali, ``Automatic and
  reliable leaf disease detection using deep learning techniques,'' {\em
  AgriEngineering}, vol.~3, no.~2, pp.~294--312, 2021.

\bibitem{selvaraju2017grad}
R.~R. Selvaraju, M.~Cogswell, A.~Das, R.~Vedantam, D.~Parikh, and D.~Batra,
  ``Grad-cam: Visual explanations from deep networks via gradient-based
  localization,'' in {\em Proceedings of the IEEE international conference on
  computer vision}, pp.~618--626, 2017.

\bibitem{gopalan2025corn}
K.~Gopalan, S.~Srinivasan, M.~Singh, S.~K. Mathivanan, and U.~Moorthy, ``Corn
  leaf disease diagnosis: enhancing accuracy with resnet152 and grad-cam for
  explainable ai,'' {\em BMC Plant Biology}, vol.~25, no.~1, p.~440, 2025.

\bibitem{preotee2024approach}
F.~F. Preotee, S.~Sarker, S.~R. Refat, T.~Muhammad, and S.~Islam, ``An approach
  towards identifying bangladeshi leaf diseases through transfer learning and
  xai,'' in {\em 2024 27th International Conference on Computer and Information
  Technology (ICCIT)}, pp.~1744--1749, IEEE, 2024.

\bibitem{muhammad2020eigen}
M.~B. Muhammad and M.~Yeasin, ``Eigen-cam: Class activation map using principal
  components,'' in {\em 2020 international joint conference on neural networks
  (IJCNN)}, pp.~1--7, IEEE, 2020.

\bibitem{jahin2025soybean}
M.~A. Jahin, S.~Shahriar, M.~F. Mridha, M.~J. Hossen, and N.~Dey, ``Soybean
  disease detection via interpretable hybrid cnn-gnn: Integrating mobilenetv2
  and graphsage with cross-modal attention,'' {\em arXiv preprint
  arXiv:2503.01284}, 2025.

\bibitem{pillay2021quantifying}
N.~Pillay, M.~Gerber, K.~Holan, S.~A. Whitham, and D.~K. Berger, ``Quantifying
  the severity of common rust in maize using mask r-cnn,'' in {\em Artificial
  Intelligence and Soft Computing: 20th International Conference, ICAISC 2021,
  Virtual Event, June 21--23, 2021, Proceedings, Part I 20}, pp.~202--213,
  Springer, 2021.

\bibitem{razzaq2019study}
T.~Razzaq, M.~F. Khan, and S.~I. Awan, ``Study of northern corn leaf blight
  (nclb) on maize (zea mays l.) genotypes and its effect on yield,'' {\em
  Sarhad J. Agric}, vol.~35, pp.~1166--1174, 2019.

\bibitem{dhami2015review}
N.~B. Dhami, S.~Kim, A.~Paudel, J.~Shrestha, and T.~R. Rijal, ``A review on
  threat of gray leaf spot disease of maize in asia,'' {\em Journal of Maize
  Research and Development}, vol.~1, no.~1, pp.~71--85, 2015.

\bibitem{sun2021southern}
Q.~Sun, L.~Li, F.~Guo, K.~Zhang, J.~Dong, Y.~Luo, and Z.~Ma, ``Southern corn
  rust caused by puccinia polysor a underw: a review,'' {\em Phytopathology
  Research}, vol.~3, no.~1, p.~25, 2021.

\bibitem{prasad2020progress}
P.~Prasad, S.~Savadi, S.~Bhardwaj, and P.~Gupta, ``The progress of leaf rust
  research in wheat,'' {\em Fungal biology}, vol.~124, no.~6, pp.~537--550,
  2020.

\bibitem{basandrai2017powdery}
A.~K. Basandrai and D.~Basandrai, ``Powdery mildew of wheat and its
  management,'' in {\em Management of wheat and barley diseases}, pp.~133--181,
  Apple Academic Press, 2017.

\bibitem{schwessinger2017fundamental}
B.~Schwessinger, ``Fundamental wheat stripe rust research in the 21st
  century,'' {\em New Phytologist}, vol.~213, no.~4, pp.~1625--1631, 2017.

\bibitem{laribi2024tan}
M.~Laribi, R.~Aboukhaddour, and S.~E. Strelkov, ``Tan spot (pyrenophora
  tritici-repentis) of wheat: A minireview,'' {\em Plant Health Cases},
  no.~2024, 2024.

\bibitem{sanya2022review}
D.~R.~A. Sanya, S.~F. Syed-Ab-Rahman, A.~Jia, D.~On{\'e}sime, K.-M. Kim, B.~C.
  Ahohuendo, and J.~R. Rohr, ``A review of approaches to control bacterial leaf
  blight in rice,'' {\em World Journal of Microbiology and Biotechnology},
  vol.~38, no.~7, p.~113, 2022.

\bibitem{fernandez2023phantom}
J.~Fernandez, ``The phantom menace: latest findings on effector biology in the
  rice blast fungus,'' {\em Abiotech}, vol.~4, no.~2, pp.~140--154, 2023.

\bibitem{chen2021approach}
S.~Chen, K.~Zhang, Y.~Zhao, Y.~Sun, W.~Ban, Y.~Chen, H.~Zhuang, X.~Zhang,
  J.~Liu, and T.~Yang, ``An approach for rice bacterial leaf streak disease
  segmentation and disease severity estimation,'' {\em Agriculture}, vol.~11,
  no.~5, p.~420, 2021.

\bibitem{surendhar2022status}
M.~Surendhar, Y.~Anbuselvam, and J.~Ivin, ``Status of rice brown spot
  (helminthosporium oryz) management in india: A review,'' {\em Agricultural
  Reviews}, vol.~43, no.~2, pp.~217--222, 2022.

\bibitem{joseph2024real}
D.~S. Joseph, P.~M. Pawar, and K.~Chakradeo, ``Real-time plant disease dataset
  development and detection of plant disease using deep learning,'' {\em IEEE
  Access}, 2024.

\bibitem{li2022survey}
Y.~Li, C.~P. Chen, and T.~Zhang, ``A survey on siamese network: Methodologies,
  applications, and opportunities,'' {\em IEEE Transactions on artificial
  intelligence}, vol.~3, no.~6, pp.~994--1014, 2022.

\bibitem{snell2017prototypical}
J.~Snell, K.~Swersky, and R.~S. Zemel, ``Prototypical networks for few-shot
  learning,'' in {\em Advances in neural information processing systems},
  pp.~4077--4087, 2017.

\bibitem{sung2018learning}
F.~Sung, Y.~Yang, L.~Zhang, T.~Xiang, P.~H. Torr, and T.~M. Hospedales,
  ``Learning to compare: Relation network for few-shot learning,'' in {\em
  Proceedings of the IEEE conference on computer vision and pattern
  recognition}, pp.~1199--1208, 2018.

\bibitem{vinyals2016matching}
O.~Vinyals, C.~Blundell, T.~Lillicrap, D.~Wierstra, {\em et~al.}, ``Matching
  networks for one shot learning,'' {\em Advances in neural information
  processing systems}, vol.~29, 2016.

\bibitem{HussainT22}
H.~Hussain and P.~S. Tamizharasan, ``The impact of cascaded optimizations in
  {CNN} models and end-device deployment,'' in {\em Proceedings of the 20th
  {ACM} Conference on Embedded Networked Sensor Systems, SenSys 2022, Boston,
  Massachusetts, November 6-9, 2022} (J.~Gummeson, S.~I. Lee, J.~Gao, and
  G.~Xing, eds.), pp.~954--961, {ACM}, 2022.

\bibitem{sy}
K.~Syama, J.~A.~A. Jothi, and N.~Khanna, ``Automatic disease prediction from
  human gut metagenomic data using boosting graphsage,'' {\em {BMC}
  Bioinformatics}, vol.~24, no.~1, p.~126, 2023.

\bibitem{mahadik2023efficient}
S.~Mahadik, P.~M. Pawar, and R.~Muthalagu, ``Efficient intelligent intrusion
  detection system for heterogeneous internet of things (hetiot),'' {\em
  Journal of Network and Systems Management}, vol.~31, no.~1, p.~2, 2023.

\bibitem{chen2021meta}
L.~Chen, X.~Cui, and W.~Li, ``Meta-learning for few-shot plant disease
  detection,'' {\em Foods}, vol.~10, no.~10, p.~2441, 2021.

\bibitem{mu2024few}
J.~Mu, Q.~Feng, J.~Yang, J.~Zhang, and S.~Yang, ``Few-shot disease recognition
  algorithm based on supervised contrastive learning,'' {\em Frontiers in Plant
  Science}, vol.~15, p.~1341831, 2024.

\bibitem{hughes2015open}
D.~Hughes, M.~Salath{\'e}, {\em et~al.}, ``An open access repository of images
  on plant health to enable the development of mobile disease diagnostics,''
  {\em arXiv preprint arXiv:1511.08060}, 2015.

\bibitem{saad2024plant}
M.~H. Saad and A.~E. Salman, ``A plant disease classification using one-shot
  learning technique with field images,'' {\em Multimedia Tools and
  Applications}, vol.~83, no.~20, pp.~58935--58960, 2024.

\bibitem{pan2022automatic}
J.~Pan, L.~Xia, Q.~Wu, Y.~Guo, Y.~Chen, and X.~Tian, ``Automatic strawberry
  leaf scorch severity estimation via faster r-cnn and few-shot learning,''
  {\em Ecological Informatics}, vol.~70, p.~101706, 2022.

\bibitem{paul2024study}
H.~Paul, S.~Ghatak, S.~Chakraborty, S.~K. Pandey, L.~Dey, D.~Show, and
  S.~Maity, ``A study and comparison of deep learning based potato leaf disease
  detection and classification techniques using explainable ai,'' {\em
  Multimedia Tools and Applications}, vol.~83, no.~14, pp.~42485--42518, 2024.

\bibitem{arvind2021deep}
C.~Arvind, A.~Totla, T.~Jain, N.~Sinha, R.~Jyothi, K.~Aditya, M.~Farhan,
  G.~Sumukh, G.~Ak, {\em et~al.}, ``Deep learning based plant disease
  classification with explainable ai and mitigation recommendation,'' in {\em
  2021 IEEE Symposium Series on Computational Intelligence (SSCI)}, pp.~01--08,
  IEEE, 2021.

\bibitem{aldakheel2024detection}
E.~A. Aldakheel, M.~Zakariah, and A.~H. Alabdalall, ``Detection and
  identification of plant leaf diseases using yolov4,'' {\em Frontiers in Plant
  Science}, vol.~15, p.~1355941, 2024.

\bibitem{jiang2022review}
P.~Jiang, D.~Ergu, F.~Liu, Y.~Cai, and B.~Ma, ``A review of yolo algorithm
  developments,'' {\em Procedia computer science}, vol.~199, pp.~1066--1073,
  2022.

\bibitem{thakur2022explainable}
P.~S. Thakur, P.~Khanna, T.~Sheorey, and A.~Ojha, ``Explainable vision
  transformer enabled convolutional neural network for plant disease
  identification: Plantxvit,'' {\em arXiv preprint arXiv:2207.07919}, 2022.

\bibitem{ghosh2023recognition}
P.~Ghosh, A.~K. Mondal, S.~Chatterjee, M.~Masud, H.~Meshref, and A.~K. Bairagi,
  ``Recognition of sunflower diseases using hybrid deep learning and its
  explainability with ai,'' {\em Mathematics}, vol.~11, no.~10, p.~2241, 2023.

\bibitem{he2016deep}
K.~He, X.~Zhang, S.~Ren, and J.~Sun, ``Deep residual learning for image
  recognition,'' in {\em Proceedings of the IEEE conference on computer vision
  and pattern recognition}, pp.~770--778, 2016.

\bibitem{chen2019closer}
W.-Y. Chen, Y.-C. Liu, Z.~Kira, Y.-C.~F. Wang, and J.-B. Huang, ``A closer look
  at few-shot classification,'' {\em arXiv preprint arXiv:1904.04232}, 2019.

\bibitem{zhe2019directional}
X.~Zhe, S.~Chen, and H.~Yan, ``Directional statistics-based deep metric
  learning for image classification and retrieval,'' {\em Pattern Recognition},
  vol.~93, pp.~113--123, 2019.

\bibitem{pedregosa2011scikit}
F.~Pedregosa, G.~Varoquaux, A.~Gramfort, V.~Michel, B.~Thirion, O.~Grisel,
  M.~Blondel, P.~Prettenhofer, R.~Weiss, V.~Dubourg, {\em et~al.},
  ``Scikit-learn: Machine learning in python,'' {\em the Journal of machine
  Learning research}, vol.~12, pp.~2825--2830, 2011.

\bibitem{koch2015siamese}
G.~Koch, R.~Zemel, R.~Salakhutdinov, {\em et~al.}, ``Siamese neural networks
  for one-shot image recognition,'' in {\em ICML deep learning workshop},
  vol.~2, pp.~1--30, Lille, 2015.

\bibitem{chattopadhay2018grad}
A.~Chattopadhay, A.~Sarkar, P.~Howlader, and V.~N. Balasubramanian,
  ``Grad-cam++: Generalized gradient-based visual explanations for deep
  convolutional networks,'' in {\em Proceedings of the IEEE Winter Conference
  on Applications of Computer Vision (WACV)}, pp.~839--847, IEEE, 2018.

\bibitem{bany2021eigen}
M.~Bany~Muhammad and M.~Yeasin, ``Eigen-cam: Visual explanations for deep
  convolutional neural networks,'' {\em SN Computer Science}, vol.~2, no.~1,
  p.~47, 2021.

\bibitem{robertson2014nclb}
A.~Robertson, ``Northern leaf blight prevalent in iowa,'' 2014.
\newblock Photo: Large elliptical lesions of Northern corn leaf blight. Image
  credited to Iowa State University Extension and Outreach. Article may be
  republished as written with credit to the author and institution.

\bibitem{isakeit2021_nclb}
T.~Isakeit and T.~A. A.~E. Service, ``Northern corn leaf blight lesions
  (figure 1),'' 2021.
\newblock Figure 1: Lesions of northern corn leaf blight caused by the fungus
  *Exserohilum turcicum*. Courtesy of Texas A\&M AgriLife Extension. All
  rights reserved.

\bibitem{jackson2012maize}
G.~Jackson, ``Setosphaeria turcica (maize leaf blight).'' Photograph, 2012.
\newblock Photo 2. Information from CABI (2012) Crop Protection Compendium.
  Available: \url{https://www.cabi.org/cpc/}. Original source: Kohler F.,
  Pellegrin F., Jackson G., McKenzie E. (1997). \textit{Diseases of cultivated
  crops in Pacific Island countries}. South Pacific Commission.

\bibitem{malvick_nclb_image}
D.~Malvick, ``Northern corn leaf blight lesion,'' 2018.
\newblock Image showing a Northern Corn Leaf Blight lesion on a corn leaf.
  Courtesy of University of Minnesota Extension. © Regents of the University
  of Minnesota. All rights reserved.

\bibitem{ocj2014corn}
{Ohio's Country Journal}, ``Disease pressures in some ohio corn fields hitting
  economic thresholds,'' July 2014.
\newblock [Accessed: 16-Jul-2025].

\end{thebibliography}

\begin{comment}

\end{comment}
\end{document}